\documentclass[10pt,letterpaper]{article}

\usepackage[letterpaper,margin=1in]{geometry} % standard margins for single column

\usepackage{newtxtext,newtxmath}   % Times/Helvetica/Courier + fine math
\usepackage[T1]{fontenc}
\usepackage[utf8]{inputenc}
\usepackage[activate=true]{microtype} % kerning, protrusion

\usepackage{graphicx}
\usepackage{subcaption}
\usepackage{booktabs}
\usepackage{siunitx}
\usepackage{float} 

\usepackage{amsmath}
\usepackage{amssymb}
\usepackage{algorithm}
\usepackage{algpseudocode}

\usepackage[font={small,sf},labelfont={bf,sf}]{caption}
\usepackage{titlesec}
\titleformat{\section}
  {\Large\sffamily\bfseries\color{brandgreen}}
  {\thesection}{1em}{}
\titlespacing*{\section}{0pt}{3.5ex plus 1ex minus .2ex}{2.3ex plus .2ex}

\titleformat{\subsection}
  {\large\sffamily\bfseries\color{brandgreen}}
  {\thesubsection}{1em}{}
\titlespacing*{\subsection}{0pt}{3.25ex plus 1ex minus .2ex}{1.5ex plus .2ex}

\titleformat{\subsubsection}
  {\normalsize\sffamily\bfseries\color{brandgreen}}
  {\thesubsubsection}{1em}{}
\titlespacing*{\subsubsection}{0pt}{3.25ex plus 1ex minus .2ex}{1.5ex plus .2ex}

\DeclareTextFontCommand{\textbf}{\bfseries\sffamily}

\usepackage[numbers]{natbib} % or numbers if you prefer
\usepackage[colorlinks=true,
            linkcolor=blue,
            citecolor=blue,
            urlcolor=blue]{hyperref}

\usepackage[capitalize,noabbrev]{cleveref}

\usepackage[dvipsnames]{xcolor}      % for \definecolor and named colors
\usepackage[most]{tcolorbox}         % provides \newtcolorbox + gradient/skins
\definecolor{brandgreen}{HTML}{1EB980}
\definecolor{brandpurple}{HTML}{6C63FF}
\definecolor{seedblue}{HTML}{2E5AA8}   % blue color for sections like in paper.tex

\definecolor{top1color}{HTML}{63D3B0}  % top-1: brandgreen
\definecolor{top2color}{HTML}{D6D4FF}  % top-2: lighter
\definecolor{top3color}{HTML}{EBF4FF}  % top-3: even lighter

\hypersetup{
  linkcolor=brandgreen,
  citecolor=brandgreen,
  urlcolor=brandgreen
}

\definecolor{color1of12}{RGB}{255, 255, 255}
\definecolor{color2of12}{RGB}{255, 255, 255}
\definecolor{color3of12}{RGB}{255, 255, 255}
\definecolor{color4of12}{RGB}{255, 255, 255}
\definecolor{color5of12}{RGB}{255, 255, 255}
\definecolor{color6of12}{RGB}{255, 255, 255}
\definecolor{color7of12}{RGB}{255, 255, 255}
\definecolor{color8of12}{RGB}{255, 255, 255}
\definecolor{color9of12}{RGB}{255, 255, 255}
\definecolor{color10of12}{HTML}{EBF4FF}
\definecolor{color11of12}{HTML}{D6D4FF}
\definecolor{color12of12}{HTML}{63D3B0}

\newtcolorbox{herobox}[1][]{enhanced,
  boxrule=1.2pt,
  arc=3mm,
  left=6mm,right=6mm,top=6mm,bottom=8mm,
  coltitle=black,
  interior style={left color=brandgreen!8, right color=brandpurple!8},
  frame style={left color=brandgreen!80!black, right color=brandpurple!80!black},
  #1}

\newcommand{\herotitle}[5]{%
  \begin{herobox}
    {\Large\bfseries #1}\\[2mm]
    {\small #2}\\[1mm]

    \noindent
    \begin{minipage}[t]{0.5\linewidth}
      {%
        \fontsize{7.8pt}{8.4pt}\selectfont
        #3% левый столбец: абстракт + Sber + Date
      }%
    \end{minipage}%
    \hfill
    \begin{minipage}[t]{0.45\linewidth}
      \centering
      \vspace*{-0.7\baselineskip}%
      \includegraphics[width=\linewidth]{#4}\\[0.8em]
      \raggedright
      \footnotesize
      #5 % правый столбец под картинкой: Code + Project Page
    \end{minipage}
  \end{herobox}%
}

\usepackage{minitoc}
\usepackage{url}
\usepackage{amssymb}
\usepackage[utf8]{inputenc}
\usepackage{microtype}
\usepackage{booktabs}
\usepackage{pifont} 
\usepackage{multirow}
\usepackage{makecell}
\usepackage{paralist}
\usepackage{xspace}
\usepackage{color}
\usepackage{xcolor}
\usepackage{colortbl}
\usepackage{adjustbox}
\usepackage{hyperref} 
\usepackage[edges]{forest}
\usepackage{tikz} 
\usepackage{caption}
\usepackage{amsfonts}
\usepackage[toc,page,header]{appendix}
\usepackage[utf8]{inputenc}
\usepackage{bbm}
\usepackage{bm}
\usepackage{enumitem}

\definecolor{seedc}{RGB}{7, 92, 173}

\newcommand{\name}[1]{}
\newcommand{\hardware}[1]{}

\renewcommand{\paragraph}[1]{\vspace{0.1em}\noindent\textbf{#1}}

\usepackage{enumitem}
\setlist{nosep,leftmargin=*,itemsep=1pt,topsep=2pt}

\newcommand{\dataqa}{\textsc{DataQA}}

\newcommand{\GreenVLA}{%
  \textcolor{brandgreen}{Green}\textcolor{brandpurple}{-VLA}%
}

\begin{document}

% ---------- HERO TITLE ----------
\begin{center}
\herotitle
  {\GreenVLA:\\[0.5em] Staged Vision–Language–Action Model for Generalist Robots}
  {Manipulation Team, Sber Robotics Center* 
  
  {\scriptsize *A detailed list of contributors in \autoref{sec:contributors}}}
{\quad We introduce \emph{Green-VLA}, a staged Vision--Language--Action framework for real-world deployment on the humanoid \emph{Green} robot, while maintaining generalization across diverse embodiments. Green-VLA follows a five-stage curriculum: (L0) foundational VLMs, (L1) multimodal grounding, (R0) multi-embodiment pretraining, (R1) embodiment-specific adaptation, and (R2) RL-based policy alignment. Progression builds semantic and physical priors, learns shared affordances, and aligns policies for long-horizon execution beyond behavior cloning. At its core is a unified data and control stack for robot fleets. A scalable data-processing pipeline including \dataqa{} and temporal-alignment filters and synchronizes 3{,}000 hours of demonstrations; a unified, embodiment-aware action interface enables a single policy to control humanoids, mobile manipulators, and fixed-base arms; and the VLA controller is enhanced with episode-progress prediction, out-of-distribution detection, and a joint-prediction-based guidance module that generalizes to unseen objects. Optimized for the Green humanoid, Green-VLA generalizes in a zero-shot manner to new embodiments and achieves state-of-the-art performance across bimanual systems and benchmarks, with RL alignment providing gains in success rate, robustness, and long-horizon efficiency.\\[1em]

    \textbf{Sber Robotics Center}\\
    Date: February 2026%

  }
  {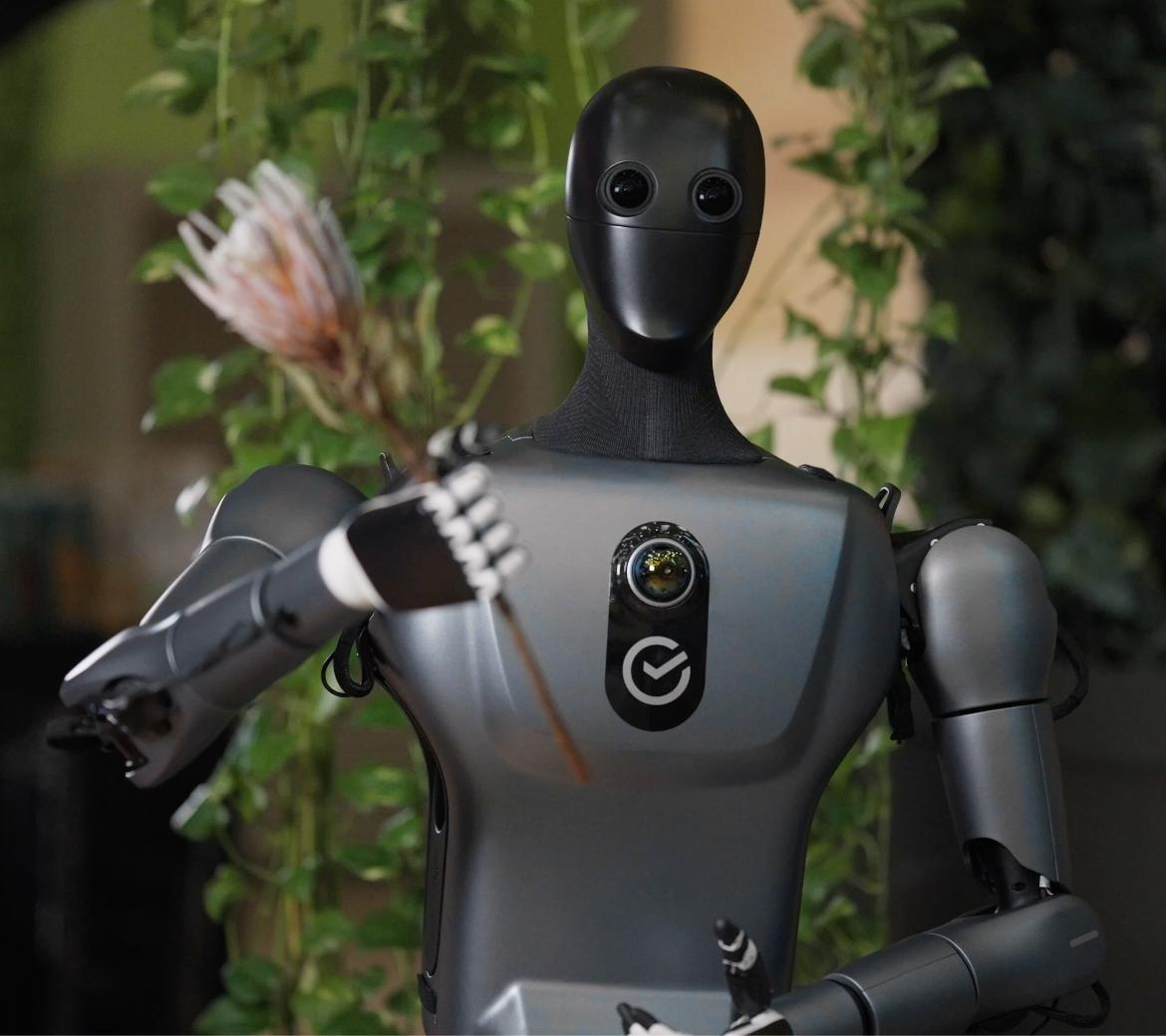}
    % {\\[0.9em]
    {
    Code: \url{https://github.com/greenvla/GreenVLA}\\
    Project Page: \url{https://greenvla.github.io/}%
  }
\end{center}

%\clearpage        
%\phantomsection   
%\tableofcontents
%\clearpage 

\section{Introduction}

Vision--Language--Action (VLA) models have recently emerged as a promising foundation for embodied AI, extending the success of large-scale language and vision models into robotics. By conditioning robot control on multimodal context and natural language instructions, VLAs aim to unify perception, reasoning, and action within a single end-to-end framework. This paradigm promises generalist robots capable of executing diverse, long-horizon tasks across heterogeneous environments, with the flexibility to adapt to novel objects, goals, and embodiments. Recent works such as $\pi_0$ \cite{pi0}, Gemini Robotics \cite{geminirobotics15}, GR00T~N1 \cite{gr00t_n1}, and AgiBot~GO-1 \cite{agibot_go1} highlight this trajectory, combining large-scale data aggregation with unified architectures and demonstrating strong progress on manipulation, reasoning, and evaluation benchmarks.

Despite this rapid progress, scaling alone does not resolve the core challenges of real-world deployment. First, robotic datasets are inherently \emph{heterogeneous} in terms of observations, action spaces, and sampling rates. Second, data quality varies drastically across sources, with trajectories suffering from jitter, blurry frames, inconsistent execution, and low scene diversity. Third, the predominant training paradigm remains behavior cloning (BC), which minimizes
\begin{equation*} \label{eq:bc-loss}
\mathcal{L}_{\mathrm{BC}} = \mathbb{E}_{(s,a) \sim \mathcal{D}} \big[ \lVert \pi_\theta(s) - a \rVert^2 \big],
\end{equation*}
where $s$ denotes the state, $a$ the action, $\pi_\theta$ the policy for robotics control and $\mathcal{D}$ the demonstration dataset, but this approach quickly saturates and fails to align policies to long-horizon objectives and task-level rewards. These limitations yield brittle models that generalize poorly across embodiments and environments, undermining the promise of scalable robotics foundation models.

In parallel, a growing body of work has begun to explore explicit reasoning within VLA models (e.g., EO-1 \cite{eo1}, WALL-OSS \cite{WALL-OSS}), showing that integrating high-level decomposition or chain-of-thought reasoning improves long-horizon planning. Yet such approaches often rely on autoregressive reasoning loops that incur significant inference latency, preventing their use in real-time robotic control. In practice, system efficiency depends not only on success rate (SR) but also on throughput, where the time-to-completion of each task compounds over extended workflows.

\begin{figure*}[t]
    \centering
    \includegraphics[width=0.95\textwidth]{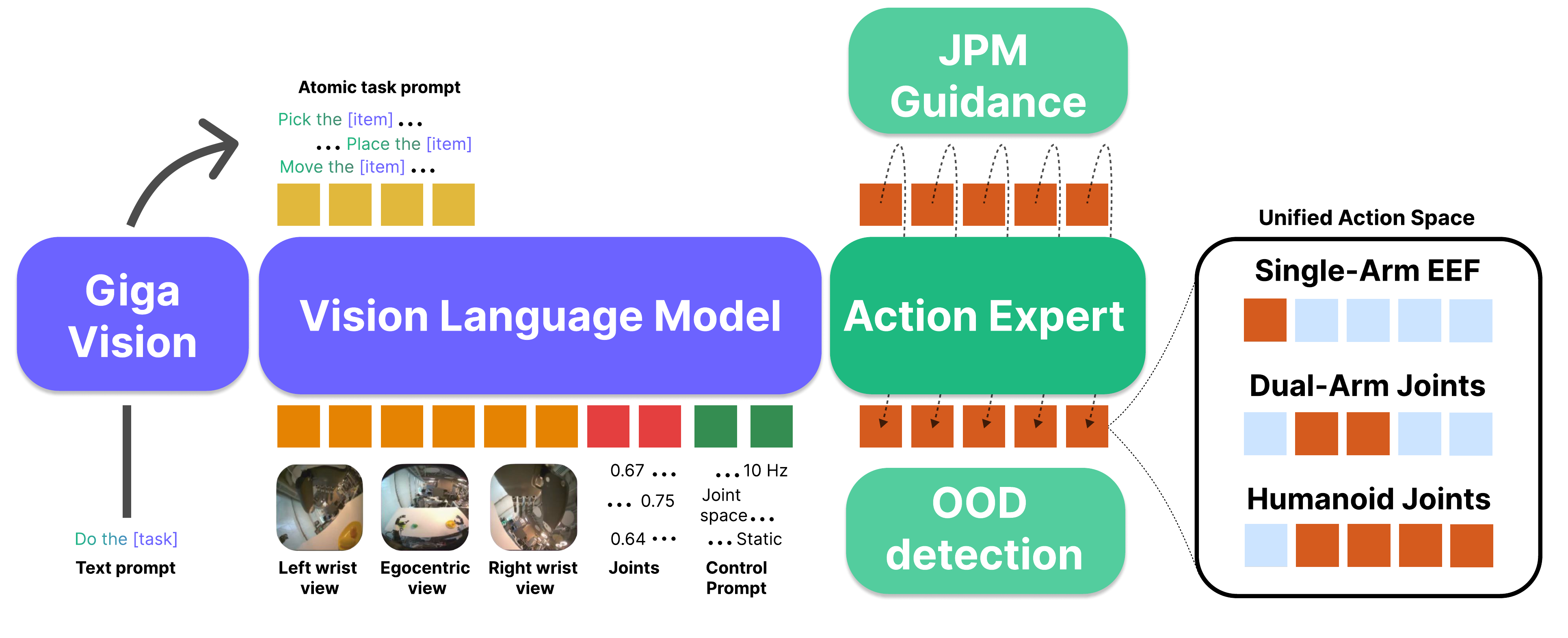}
    \caption{Green-VLA architecture. A multimodal vision–language model encodes instructions, camera views, and proprioception into tokens that feed a flow-matching action expert. A high-level task planner decomposes user goals into subtasks, queries the VLA loop, and uses auxiliary signals (episode end, OOD, and JPM-based guidance for precise target points) to ensure safe, instruction-faithful execution across embodiments.}
    \label{fig:architecture-full}
\end{figure*}

We introduce \textbf{Green-VLA}, a framework that moves \emph{beyond data scaling} by emphasizing \emph{quality alignment}, \emph{action unification}, and \emph{reinforcement learning refinement}. Green-VLA unifies over 24M non-robotics, internet-scale multimodal samples and $3{,}000$ hours of humanoid and manipulator demonstrations into a normalized action space $\mathcal{A}_u$, achieved through frequency interpolation/extrapolation and explicit control-type prompting. A \dataqa{} pipeline automatically evaluates and filters episodes using trajectory smoothness $J$, image sharpness $S$, visual diversity $D$, and state variance $\sigma^2$, with trajectory smoothing applied to reduce high-frequency noise. This ensures that scale is complemented by quality, yielding a more stable foundation for generalist learning.  

Green-VLA supports multiple embodiments and control types under the same semantic layout, enabling positive transfer between humanoids, mobile manipulators, and collaborative arms. Combined with a staged training recipe-from web-scale VLM pretraining through robotics pretraining, embodiment specialization, and RL refinement—this pipeline turns heterogeneous data into a single, consistent policy that can be deployed across diverse real-world robots. 

While the framework is embodiment-agnostic, our primary target platform is the \emph{Green} humanoid robot, where we control 32 DoF of the full upper body (head, torso, dual arms, and dexterous hands) through the unified action interface. This setting requires coordinated bimanual manipulation, whole upper-body motion, and fine-grained fingertip control, making it substantially harder than standard single-arm, parallel-gripper benchmarks. Despite this, the same Green-VLA policy successfully handles the full evaluation scope—from single-arm manipulators and dual-arm platforms to the Green humanoid—without architectural changes. This demonstrates that unified actions plus staged training are sufficient to bridge from conventional manipulators to high-DoF humanoid control.

Training proceeds in five progressive stages:
\begin{equation*} \label{eq:training-stages}
\begin{array}{c} 
L0: Base \ VLM\\
\Downarrow \\[2pt]
L1: Web \ Pretrain \ for \ Physical \ World \ Understanding\\
\Downarrow \\[2pt]
R0: General \ Robotics \ Pretrain \\
\Downarrow \\[2pt]
R1: Embodiment \ SFT \\
\Downarrow \\[2pt]
R2: RL \ Alignment
\end{array}
\end{equation*}
This recipe first leverages large-scale multimodal data for common-sense grounding, then transfers to unified robot data, adapts to specific embodiments, and finally aligns policies with reinforcement learning objectives to improve long-horizon control.

Our contributions are as follows:
\begin{enumerate}
    \item A quality and temporal alignment module for diverse robotics datasets. A \dataqa{} pipeline (jitter $J$, sharpness $S$, diversity $D$, variance $\sigma^2$) with trajectory smoothing and optical flow–based speed alignment, combined with a balanced-to-target sampler $w_i(\alpha)$ and a simple speed-conditioned modulation for multi-scale (fast/precise) control. On top of this, a joint prediction + guidance module (JPM) improving precise object targeting, especially in visually dense setups such as e-commerce shelves.
    
    \item A staged VLA training recipe bridging web-scale priors, robotics data, embodiment specialization, and RL alignment (L0$\rightarrow$L1$\rightarrow$R0$\rightarrow$R1$\rightarrow$R2), providing a clear path from generic multimodal pretraining to real-world robot deployment.

    \item Validation of the staged recipe across phases and embodiments. Green-VLA matches or exceeds prior pretrains at R0, is competitive with other VLA models after embodiment tuning, and achieves the largest gains after RL alignment (R2), especially on long-horizon success, recovery, and precise task following—showing that careful unification, data curation, and guidance matter for high-quality VLA.

    \item A deployment-ready design for the Green humanoid robot, with unified upper-body control (arms, hands, head, torso) and task-planner integration, while remaining compatible with a wide spectrum of other embodiments and standard simulators.
\end{enumerate}

\noindent
\textbf{Future work.} To fully realize the potential of Green-VLA, several extensions are envisioned. First, incorporating multilingual instruction following (English, Russian, and others) will improve inclusivity and data efficiency in global deployments. Second, adding a lightweight reasoning module for task decomposition, while preserving low-latency control, can combine the strengths of chain-of-thought planning with real-time execution. Third, integrating embodied memory and trajectory replay may further improve performance in long-horizon household or industrial tasks. Together, these directions highlight a path toward practical, scalable, and generalist robotic intelligence.

\section{Why a Staged VLA Pipeline Matters}

Scaling Vision--Language--Action (VLA) models is not only a question of parameter count or dataset size. 
Evidence from recent works --- $\pi_{0.5}$ \cite{pi05}, EO-1 \cite{eo1}, WALL-OSS \cite{WALL-OSS}, and Gemini Robotics \cite{deepmind2025} --- suggests that strong performance depends on combining complementary data regimes: web-scale multimodal data for semantic and physical common sense, and large-scale robotics action data for grounded control.
We argue that staged training is essential to balance generalization, efficiency, and real-world reliability.

\begin{itemize}[leftmargin=*, itemsep=3pt]
  \item \textbf{Base VLM (L0)} is the underlying vision–language model we start from, already pretrained on large-scale image/video–text data. It has no robot actions yet and lacks the refined visual, physical, multi-view understanding that is crucial for robots; later stages (L1–R2) adapt it to real-world execution.
  \item \textbf{Web and multimodal pretraining (L1)} builds general reasoning and semantic grounding. 
  Models exposed to internet-scale video and multimodal corpora acquire priors about physics, object affordances, and task structure 
  that cannot be recovered from robot data alone. 
  EO-1 \cite{eo1} demonstrated that adding L1-like knowledge improves zero-shot generalization to unseen objects and scenes, 
  while WALL-OSS \cite{WALL-OSS} showed that multimodal co-training with chain-of-thought signals benefits reasoning-intensive tasks. 
  In Green-VLA we use multimodal web data for L1 pretraining.

  \item \textbf{General robotics pretraining (R0)} captures broad affordance priors: mapping goals, objects, and kinematics to feasible action distributions. 
  Large-scale multi-embodiment data (humanoids, manipulators) encourage models to abstract away embodiment-specific quirks and learn cross-domain invariants. 
  $\pi_{0.5}$ showed that cross-source pretraining was crucial for achieving long-horizon execution in unseen homes. 
  Similarly, AgiBot~GO-1’s latent planner leveraged scale to gain robustness in dual-arm dexterity. 
  In Green-VLA, R0 serves as the core repository of base manipulation skills, maximizing data efficiency across embodiments.

  \item \textbf{Effective tuning (R1)} converts capacity into competence. 
  After general pretraining, careful adaptation to a target embodiment yields immediate success rate (SR) gains without requiring new large-scale data. 
  Key techniques include: 
  (i) targeted hyperparameter and optimizer search; 
  (ii) architectural adjustments such as embodiment-aware state/action heads; 
  (iii) efficiency improvements for inference (e.g., SDPA attention \cite{litman2025scaleddotproductattentiononesidedentropic} kernels, reducing denoising steps in flow matching). 
  In WALL-OSS \cite{WALL-OSS}, MoE layers \cite{shazeer2017outrageouslylargeneuralnetworks} and efficient inference design reduced overhead, highlighting that tuning must balance success rate (SR) improvements with real-time deployability.

  \item \textbf{RL alignment (R2)} closes the last-mile gap. 
  Behavior cloning saturates quickly in long-horizon and contact-rich manipulation, as it struggles to assign credit over extended action chains. The notorious problem of out-of-distribution (OOD) actions is also difficult to mitigate through additional demonstrations alone, due to the cost of human labor and the complexity of predicting OOD states.
  Reinforcement learning methods reshape the objective to incorporate task rewards, failure recovery, and preference-like feedback. 
  This stage improves both success rate and average chain length (ACL). 
  To improve performance on hard, dexterous tasks, RL-style fine-tuning is required. In Green-VLA, R2 integrates BC priors with RL alignment, achieving both stability and long-horizon robustness.

\end{itemize}

\begin{figure*}[ht]
    \centering
    \includegraphics[width=0.9\textwidth]{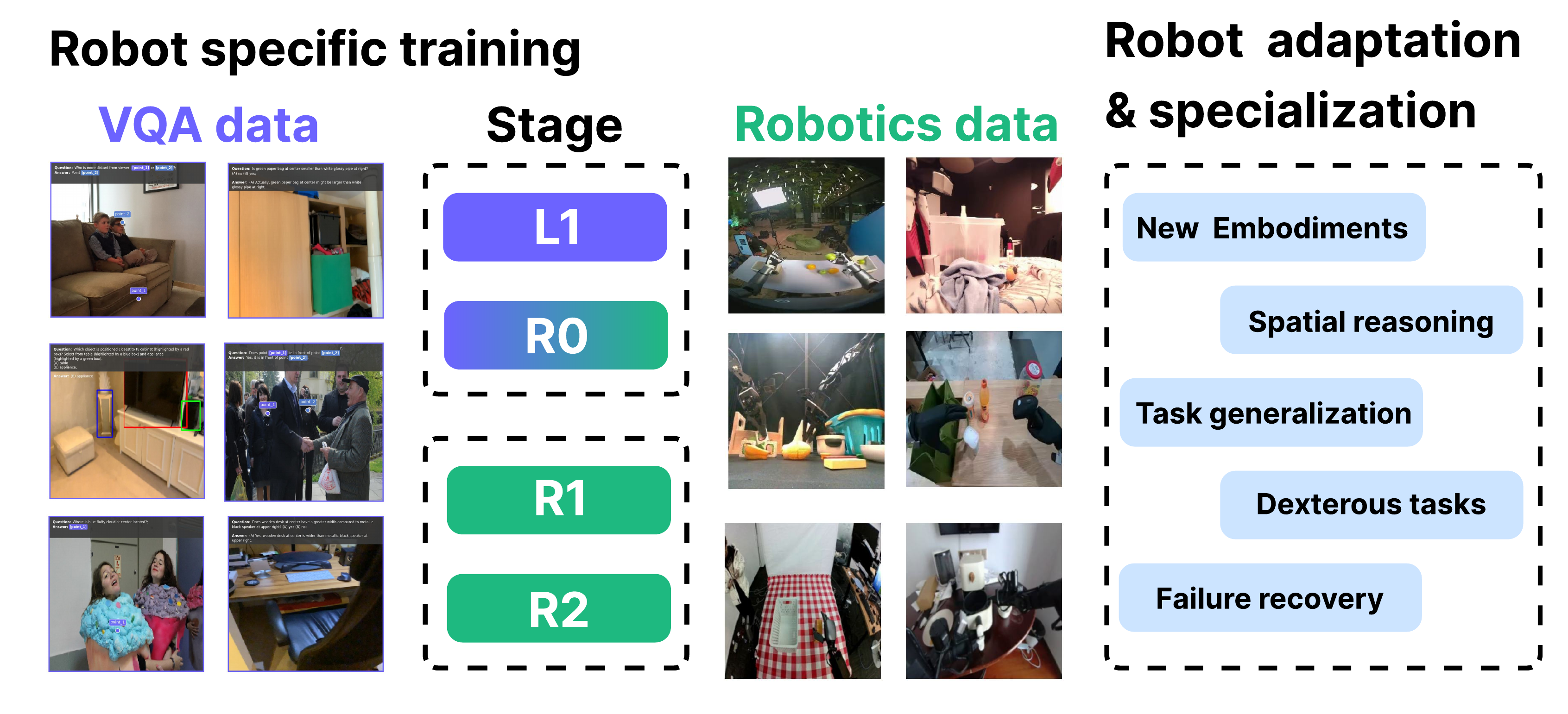}
    \caption{Green-VLA’s robot-specific training stages use visual question answering (VQA) and robotics data and enable robot adaptation and specialization for new embodiments, spatial reasoning, task generalization, dexterous manipulation, and failure recovery.}
    \label{fig:stage_specialization}
\end{figure*}

\noindent
In summary, the staged pipeline matters because each stage addresses a distinct bottleneck: 
L1 enriches semantic grounding, R0 captures affordance priors, R1 adapts efficiently to embodiments, and R2 injects reward-based alignment for real-world robustness.

\section{Green-VLA Data Framework}

\subsection{Data Pipeline}
\begin{figure*}[ht]
    \centering
    \includegraphics[width=0.8\textwidth]{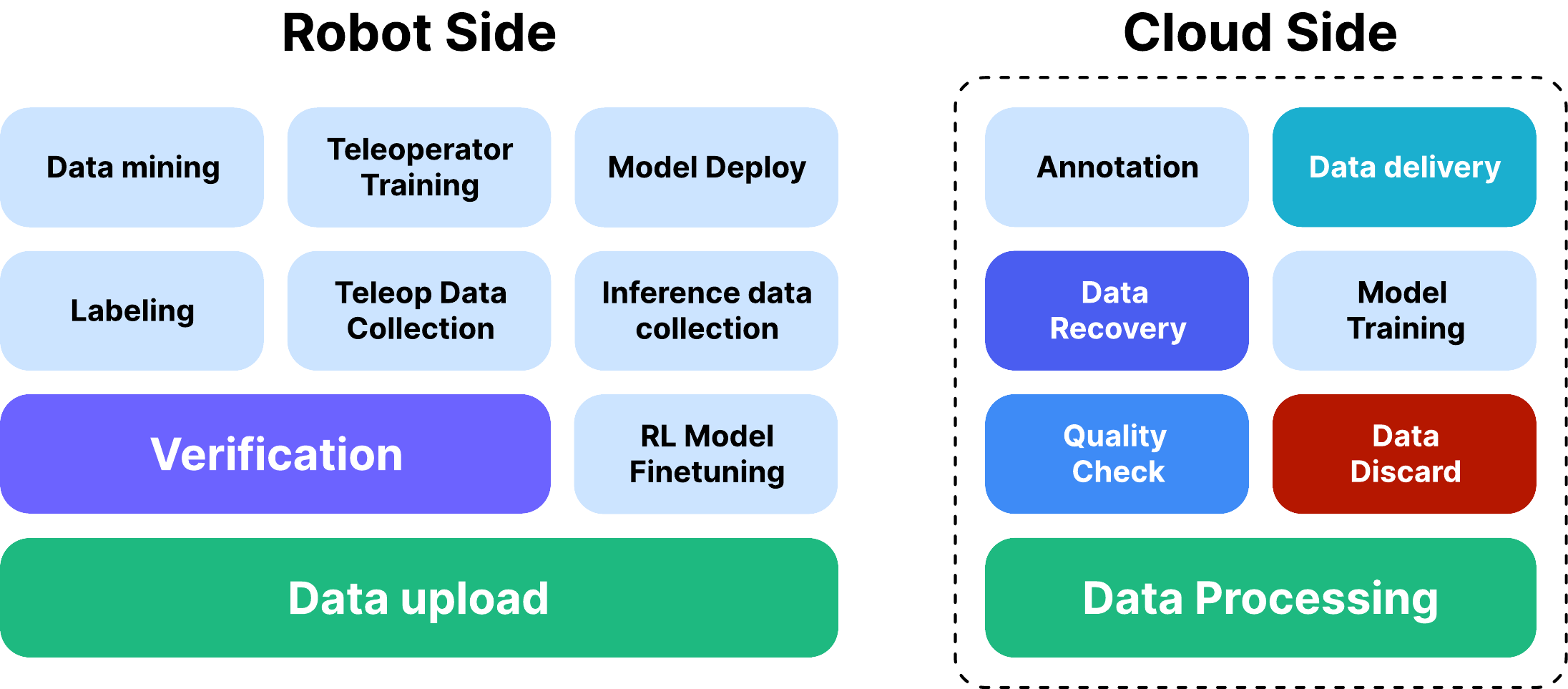}
    \caption{Overview of the data pipeline for robot learning. Data collection and processing loop integrating robot-side teleoperation, cloud-based data verification, open-source dataset mining, and model training. The pipeline supports iterative model updates via RL fine-tuning and feedback from real-robot deployments.}
    \label{fig:data_pipeline}
\end{figure*}

We train Green-VLA in stages, combining web/multimodal grounding with large-scale robotics data.
For L1 we use 24M non-robotics, internet-scale multimodal samples to learn general visual–language priors with multi-view, physical, and environment understanding. For R0, we pretrain on 184M robotics-domain samples overall $>3{,}000$ hours across humanoids and manipulators from web and our data collection pipeline. Each episode is annotated with a language instruction. RGB streams and proprioception are temporally normalized so that comparable physical progress aligns across sources. We have \dataqa{} pipeline that scores and filters trajectories using jitter $J$, diversity $D$, sharpness $S$, and state variance $\sigma^2$, discarding low-quality segments. The result is a high-quality, embodiment-agnostic corpus that preserves semantic consistency while enabling scalable, unified pretraining. An overall representation of our data collection, filtering and storage pipeline is presented in \autoref{fig:data_pipeline}.

\subsection{Dataset}
\subsubsection{Multi-modal Web Data}
\begin{figure*}[ht]
    \centering
    \includegraphics[width=\textwidth]{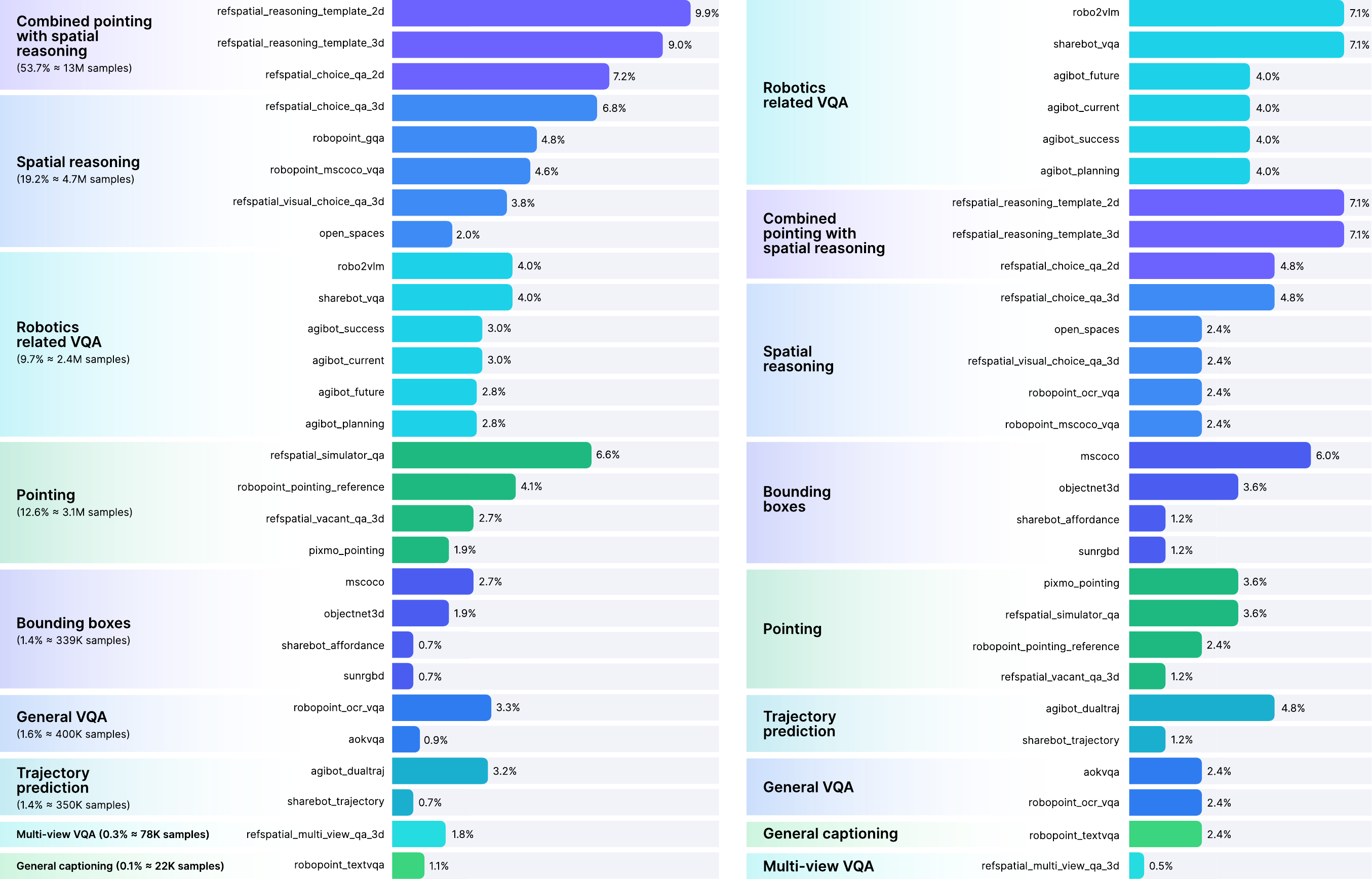}
    \caption{Datasets mixture used in L1 training phase. \textit{Left}: distribution of sample counts across sub-datasets. 
    \textit{Right}: sampling weight allocation across categories. 
    The data corpus integrates diverse web sources covering spatial reasoning, pointing, robotics-related VQA, and multi-view QA.}
    \label{fig:web-data}
\end{figure*}

For the pretraining (R0) phase, in addition to robotics data we utilized a large corpus of web data (L1) (\autoref{fig:web-data}), including general VQA, pointing, bounding box prediction, pixel-wise trajectory prediction, multi-view VQA, general captioning and combined pointing with spatial reasoning. A more detailed view of the datasets is provided below:
\begin{itemize}
  \item \emph{RefSpatial} \cite{zhou2025roborefer} --  a comprehensive dataset combining 2D and 3D images, pointing, mutli-view question answering. During training, we map points to PaliGemma special tokens, so the samples combine text tokens with spatial tokens in queries and answers.
  \item \emph{AgibotWorld} \cite{agibot_go1} -- a large-scale robotics dataset. Based on the Agibot dataset markup, we sample a few images from the main camera, then using the sampled indices and the following subtask, we create VQA samples, such as predicting the next affordable subtask, task completion answering and answering which task is currently being executed. Additionally, we project end-effector poses to main camera frames, filter idle points and randomly choose 300K images for pixel-wise trajectory prediction.
  \item \emph{RoboPoint} \cite{yuan2024robopoint} -- a dataset containing 1432  image-QA instances, including object references, synthetic free space reference, object detection and general-purpose VQA instances.
  \item \emph{ShareRobot} \cite{chen2025robo2vlm} -- a high-quality dataset including task planning, object affordance, and end-effector trajectory in pixel space. The dataset was generated synthetically from OXE data.
  \item \emph{Robo2VLM} \cite{chen2025robo2vlm} -- a synthetic dataset generated using Gemini-2.5-Pro to generate reasoning traces supporting the correct choice.
  \item \emph{PixMo-Points} \cite{deitke2024molmo} -- PixMo-Points is a dataset of images paired with referring expressions and points marking the locations the expression refers to in the image. It was collected using human annotators and contains a diverse range of points and expressions, with many high-frequency (10+) expressions.
  \item \emph{MS COCO} \cite{lin2014mscoco} -- a large-scale object detection, segmentation, and captioning dataset. In our training pipeline, we used only the object detection subset.
  \item \emph{A-OKVQA} \cite{schwenk2022aokvqa} --  a crowdsourced dataset composed of a diverse set of about 25K questions requiring a broad base of commonsense and world knowledge to answer.
  \item \emph{OpenSpaces} \cite{robospaces2024openspaces} -- is a synthetic dataset specifically designed for training vision-language models on spatial visual question answering (VQA), containing approximately 3.4 million question-answer pairs derived from localized narratives and spatial reasoning tasks.
  \item \emph{Sun RGB-D} \cite{song2015sunrgbd} -- is a benchmark 3D scene understanding dataset comprising 10,335 RGB-D images (combining RGB color with depth information) annotated with 146,617 2D polygons, 64,595 precisely oriented 3D bounding boxes, and complete room layout annotations collected from four different types of RGB-D sensors.

\end{itemize}

We use only 2D subset for pretraining. For consistency between different tasks we select custom weights while sampling the data during pretraining, see \autoref{fig:web-data}.

\subsubsection{Robotics data}
\begin{figure*}[t]
    \centering
    \includegraphics[width=1.0\textwidth]{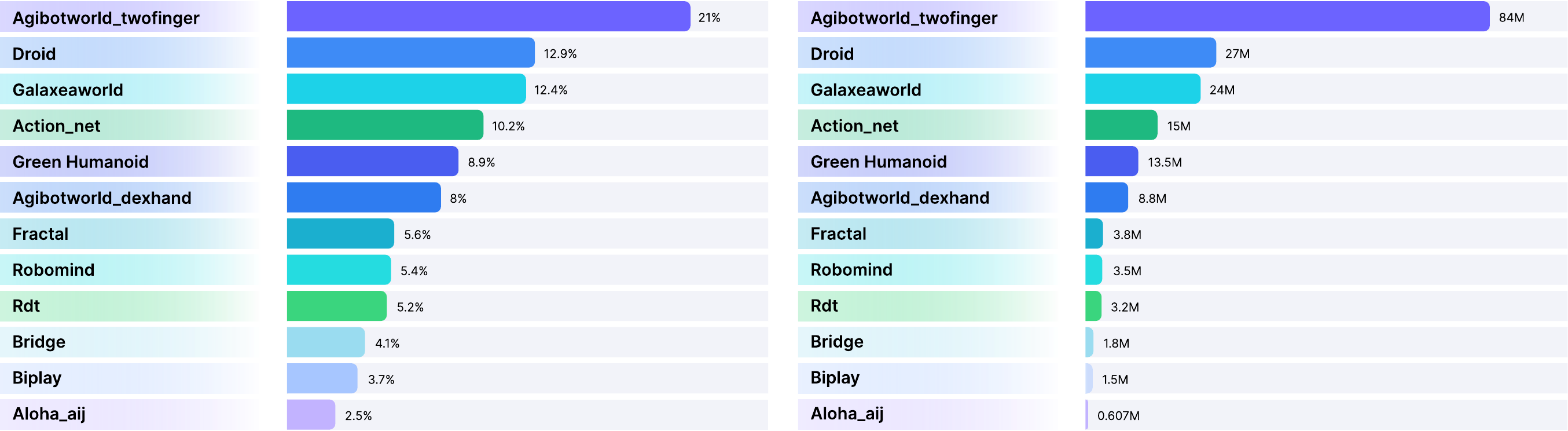}
    \caption{\textit{Left}: Dataset sampling rates used during the R0 phase of Green-VLA training. \textit{Right}: Number of data samples (frames) per dataset, illustrating relative temporal coverage. The corpus combines large-scale open datasets (e.g., AgibotWorld, DROID, Galaxea) with internally collected humanoid and dexterous-hand data.}
    \label{fig:robotics-data-summary}
\end{figure*}

% R0
% Artemiev Davidenko
For the pretraining stage (R0), we assembled a large corpus of data composed of open-source and internally collected datasets, as summarized in \autoref{fig:robotics-data-summary}.

\begin{itemize}
\item \emph{AgiBotWorld\_twofinger}\label{AgiBotWorld_twofinger} \cite{agibot_go1} features a dual-arm robot with two-finger grippers on the proprietary AgiBot A2D platform. It provides three camera views (hand\_left, hand\_right, and head) and records both Cartesian and joint states for the arms, as well as platform and torso configurations. The platform supports tilting motions, which enhance its manipulation capabilities. The dataset contains 774 hours of data, 41 unique tasks, and 18 distinct skills, primarily related to household chores and shopping activities.

\item \emph{Droid} \cite{khazatsky2024droid} is collected using a Franka Panda 7-DoF arm equipped with a Robotiq 2F-85 gripper. It uses three cameras: two static diagonal views and one wrist-mounted view. The data is recorded in Cartesian coordinates and Euler angles of the end effector during teleoperation with an Oculus Quest 2. With 49,629 unique tasks, this dataset contains 512 hours of data and 37 distinct skills, offering substantial diversity across real-world household scenarios and abstract object interactions.

\item \emph{Galaxea-Open-World-Dataset} \cite{galaxea2025} utilizes a mobile dual-arm R1 Lite manipulator with three cameras mounted on the head and wrists. It records only joint states, including platform and torso joints. The dataset comprises 477 hours of data, 9,768 unique tasks, and 85 distinct skills collected across 11 physical domains such as residential, catering, retail, and office environments.

\item \emph{Action\_net} \cite{fourier2025actionnet} collects joint-state data from humanoid Fourier robots equipped with two types of hands. Data acquisition is performed via teleoperation using Apple Vision Pro, covering 143 hours of data, 1,577 unique tasks, and 16 distinct skills.

\item \emph{AgiBotWorld\_dexhand} \cite{agibot_go1} features a robot with two five-fingered hands and provides both joint and Cartesian states. It includes three cameras (two wrist-mounted and one head-mounted) and contains 82 hours of data with 18 unique tasks and a set of distinct manipulation skills (e.g., grasping and in-hand manipulation). 

\item \emph{Fractal} \cite{rt22023arxiv} is collected on a real robot developed by Everyday Robots equipped with two-finger grippers. It employs a single camera mounted behind the manipulator at head level and records Cartesian states. The dataset contains 350 hours of data, 598 unique tasks, and 4 distinct skills.

\item \emph{Robomind} \cite{wu2025robomind} is a multi-embodiment dataset integrating data from several robots, including the Franka Emika Panda, the X-humanoid Tienkung, the Agilex cobot, and the UR5e. It uses up to three cameras and primarily features two-finger grippers, except for the humanoid robot, which has five-fingered hands. The dataset includes 33 hours of data, 389 unique tasks, and 45 distinct skills across domestic, industrial, kitchen, office, and retail domains. We used only humanoid subset from this dataset.

\item \emph{RDT} \cite{liu2024rdt} is acquired using the ALOHA dual-arm robot with two-finger grippers and three cameras positioned at chest level and on both wrists. It records joint states and comprises 60 hours of data, 271 unique tasks, and 36 distinct skills.

\item \emph{Bridge} \cite{walke2023bridgedata} is collected in real-world scenarios and involves a WidowX 250 6-DoF robot arm with a two-finger gripper. The data includes Cartesian coordinates, Euler angles, and up to four camera views, depending on the task. It provides 105 hours of data, 24 environments, 105 unique tasks, and 13 distinct skills.

\item \emph{BiPlay} \cite{dasari2025ingredients} consists of 9.7 hours of bimanual data collected with an ALOHA robot. It contains joint states, 326 unique scenes, 2,440 unique tasks, and 48 distinct skills.
\end{itemize}

There are 2 self-collected datasets in our pretraining:

\begin{itemize}
\item \emph{Green Humanoid} dataset records 32 joint states with three fisheye cameras (two wrist-mounted and one on the head) over 48 hours, encompassing 5,099 unique tasks and 4 skills.

\item \emph{ALOHA any\_pick} dataset is gathered on a modified ALOHA Agilex dual-arm platform with differential drive. It captures joint states from two wrist cameras and one platform camera, covering 11.2 hours, 1,852 unique tasks, and two unique skills, namely ``Pick <object> from the table'' and ``Place <object> into the box''.
\end{itemize}

To increase the effective amount of humanoid data, we synthetically expand the raw 48 hours of the \textit{Green Humanoid} dataset up to 167 training hours using two structured augmentations. First, we exploit the approximate bilateral symmetry of the robot to build a mirrored copy of each episode: wrist-camera streams are horizontally flipped and swapped (left $\leftrightarrow$ right), the head camera is horizontally flipped, and joint trajectories are transformed by swapping left/right limbs and negating the relevant yaw/roll components of the torso, neck, and arms so that mirrored states remain kinematically valid. Task texts are updated by swapping left/right mentions. Second, we generate time-reversed demonstrations, but only for episodes whose tasks correspond to physically reversible interactions. Concretely, we select episodes whose task templates indicate reversible skills (e.g., pick, hand-over, moving an object between hands, take from the hand), and exclude inherently irreversible ones (e.g., placing with a possible drop or irreversible release), so that the terminal state of the original trajectory is a valid initial condition for the reversed one. For a selected episode with image streams $I_t$, robot states $s_t$, and actions $a_t$ for $t = 0,\dots,T-1$, we construct a reversed episode by reordering the sequence in time,
$I'_t = I_{T-1-t}$, $s'_t = s_{T-1-t}$, and reassigning actions $a'_t$ so that each $a'_t$ drives the system from $s'_t$ to $s'_{t+1}$. The corresponding language instructions are updated to match the reversed intent (e.g., ``pick <object> from the table'' $\rightarrow$ ``place <object> on the table''). These mirrored and time-reversed trajectories are then added back to the training mixture, yielding 167 hours of effective humanoid data from 48 hours of real-world recordings.

Together, these datasets provide extensive real-world data with rich multimodal inputs and diverse task sets, offering valuable resources for developing robust bimanual manipulation policies in the R0.

% Не уверен, что тут все, что мы использовали. Например, не вижу roboset

\subsection{Data Quality Assurance}
To ensure high-quality training data, we employed the \dataqa\ pipeline for robotic dataset curation and quality assessment. This pipeline enabled us to focus model training on high-quality demonstrations, improving both sample efficiency and final policy performance.

% Кажется, что тту можно отдельным блоком фильтрацию выделить или она входит в verification?

We applied a multi-stage filtering process to remove problematic episodes from our training dataset. The filtering criteria include:
\begin{itemize}
    \item Missing cameras and missing frames.
    \item Too short and suspiciously long episodes.
    \item Motion activity thresholds to ensure meaningful robot movements.
    \item We identify erratic motions using a tremble score:
    \begin{equation*}\label{eq:Stremble}
        S_{\text{tremble}} = \frac{|\dot{s}_{\text{smooth}} - \dot{s}|}{|\dot{s}_{\text{smooth}}| + |\dot{s}|},
    \end{equation*} where $\dot{s}$ is the velocity trajectory, and $\dot{s}_{\text{smooth}}$ is its Gaussian-smoothed version.
    \item To estimate image sharpness, we first detect local regions with sharp boundaries using Laplacian std score in blocks of size $4\times4$ pixels. These local boundary scores are max-pooled with a kernel of size $16\times16$ to estimate which regions of $64\times64$ pixels are sharp and which are blurry/low-textured, i.e., do not contain sharp boundaries. The median of these region sharpness scores is considered as the episode sharpness score: $S_{\text{sharp}} = \text{median}(\text{MaxPool}(\text{std}{\text{block}}(\nabla^2 I)))$.
    \item Gripper open/closed action pattern validation. It is a task-specific filter. For example a pattern ``open-closed-open'' is expected for pick-and-place tasks.
\end{itemize}

Beyond filtering, we evaluated dataset diversity and quality using quantitative metrics to weigh datasets in the training mixture. Visual diversity was quantified through analysis of DINOv3~\cite{simeoni2025dinov3} features $f$ as 
\begin{equation*} \label{eq:Dvis}
    D_{\text{vis}} = \mathbb{E}_d[\text{std}_t(\mathbb{E}_s[f_{t,s,d}])].
\end{equation*}
% $D_{\text{vis}} = \frac{1}{D}\sum_d[\text{std}_t(\frac{1}{S}\sum_s[\mathbf{f}_{t,s,d}])]$
  Here, features $f$ are averaged over spatial locations $s$, then standard deviation is computed over time $t$, and averaged across feature dimensions $d$. State-space diversity was measured as the Frobenius norm of the covariance matrix of robot states in the dataset: $D_{\text{state}} = \sqrt{\text{tr}(\text{Cov}(s))}$, where $s$ represents the concatenated proprioceptive signals.

  Dataset quality metrics and sampling probabilities are presented in \autoref{tab:dataset_quality}.
  Outlier $D_\text{state}$ values arise from state spaces that vary in their units across embodiments. AgiBot twofinger received the highest sampling weight as it is largest (700+ hours), and is the most visually diverse among humanoid datasets. Among ALOHA datasets, Galaxea is the most visually diverse and the largest, so it received the highest weight. Galaxea contains the base movement which results in higher $D_\text{state}$ than other ALOHA datasets.  % low sharpness
  Among single-arm datasets, DROID has the highest weight as its trajectories are smoother, scenes and tasks are more diverse, and it is larger than other datasets.

\begin{table}[h!]
\centering
\caption{Dataset quality metrics and sampling probabilities.}
\label{tab:dataset_quality}
\begin{tabular}{clcccccc|c}
\toprule
&\textbf{Dataset} & \textbf{\#Episodes} & \textbf{Hours} & \textbf{$D_\text{vis}\uparrow$} & \textbf{$D_\text{state}\uparrow$} & \textbf{$S_{\text{sharp}}\uparrow$} & \textbf{$S_{\text{tremble}}\downarrow$} & \textbf{Sampling prob.}\\
\midrule
\multirow{5}{*}{\adjustbox{angle=90,origin=c}{\parbox{1.75cm}{Humanoid}}} 
& RoboMind & 6K & 33 & 0.085 & 1.72 & \cellcolor{top1color}150.00 & 0.402 & 0.054\\
& AgiBot dexhand & 6K & 82 & 0.128 & \cellcolor{top1color}4391.79 & 77.17 & \cellcolor{top3color}0.261 & 0.080\\
& AgiBot twofinger & 46K & \cellcolor{top1color}774 & 0.149 & \cellcolor{top2color}3975.09 & 83.90 & 0.286 & \cellcolor{top1color}0.210 \\
& ActionNet & 30K & 143 & 0.088 & 6.84 & \cellcolor{top2color}124.41 & 0.341 & 0.102\\
& Green Humanoid (ours) & \cellcolor{top1color}135K & 167 & 0.119 & 2.00 & 82.64 & \cellcolor{top2color}0.256 & 0.089 \\
\midrule
\multirow{4}{*}{\adjustbox{angle=90,origin=c}{\parbox{1.4cm}{ALOHA}}} 
& ALOHA any\_pick (ours) & 7K & 11 & 0.126 & 2.62 & 99.49 & 0.268 & 0.025\\
& BiPlay & 7K & 31 & 0.104 & 1.63 & 84.59 & 0.402 & 0.037\\
& RDT & 6K & 59 & \cellcolor{top2color}0.162 & 2.70 & \cellcolor{top3color}101.39 & 0.319 & 0.052\\
& Galaxea & \cellcolor{top2color}114K & \cellcolor{top3color}477 & \cellcolor{top1color}0.168 & \cellcolor{top3color}37.85 & 37.81 & 0.398 & \cellcolor{top3color}0.124\\
\midrule
\multirow{3}{*}{\adjustbox{angle=90,origin=c}{\parbox{1.05cm}{1 Arm}}} 
& Bridge & 53K & 105 & \cellcolor{top3color}0.162 & 0.69 & 41.44 & 0.358 & 0.041\\
& DROID & \cellcolor{top3color}92K & \cellcolor{top2color}501 & 0.140 & 3.10 & 60.59 & \cellcolor{top1color}0.137 & \cellcolor{top2color}0.129\\
& Fractal & 87K & 351 & 0.106 & 0.95 & 60.77 & 0.399 & 0.056\\
\bottomrule
\end{tabular}
\end{table}

\section{Green-VLA Model}
We first present the architecture of Green-VLA in \autoref{fig:architecture-full} and describe the R0 and R1 training phases. Conceptually, during R0, we train the model using all available robotics data. During R1, we fine-tune the model in the same manner, but using a high-quality dataset tailored to the target embodiment. We then describe the R2 reinforcement learning fine-tuning stage.

% \begin{figure*}[t]
%     \centering
%     \begin{subfigure}[t]{0.52\textwidth}
%         \centering
%         \includegraphics[width=\textwidth]{images/web_data/DS_left.pdf}
%         \caption{Distribution of sample counts across sub-datasets after power-scaled normalization ($\gamma = 0.4$)}
%         \label{fig:sphere}
%     \end{subfigure}
%     \hfill
%     \centering
%     \begin{subfigure}[t]{0.46\textwidth}
%         \centering
%         \includegraphics[width=\textwidth]{images/web_data/DS_right.pdf}
%         \caption{Sampling weight allocation across categories used during training.}
%         % Sampling weights in Multi-Modal web data}
%         \label{fig:sphere}
%     \end{subfigure}
%     \hfill
%     \centering
%     \caption{The dataset integrates diverse web sources covering spatial reasoning, pointing, robotics-related VQA, and multi-view QA. }
%     \label{fig:web-data}
% \end{figure*}

\subsection{Architecture}
Green-VLA builds on a unified transformer-based architecture that maps rich multimodal context to normalized actions in $\mathcal{A}_u$. A vision–language encoder first fuses RGB observations, proprioceptive state, and natural language instructions into a shared token sequence, augmented with an embodiment/control-type prompt $c$ that specifies the active effectors and action parameterization. On top of this representation, a flow-matching action expert predicts unified action chunks in $\mathcal{A}_u$, respecting our semantic slot layout and masks for each robot, while optional guidance steers trajectories toward task-specific targets. We further optimize the architecture for real-time deployment through efficient attention (SDPA kernels), lightweight heads, and reduced denoising steps, enabling a single, scalable policy that supports multi-embodiment control with low latency.

\subsection{Task planner module}
Our task planner (\autoref{fig:architecture-full} left) is a high-level GigaVision VLM  based on GigaChat \cite{mamedov2025gigachat} that sits on top of Green-VLA and decides when and how the low-level policy should act. The planner first parses voice or text input and classifies whether the query requires physical action. If no action is needed (e.g., a question about the environment or system status), high-level VLM is used purely as a language model and replies via text or speech. Otherwise, the planner converts the user’s high-level goal (e.g., “set the table for lunch”, “prepare an order for pickup”) into an intermediate, symbolic code: a sequence of atomic subtasks such as pick [item] with [left/right] hand, place [item] on [target], give [item]. This code is then translated into a structured prompt that conditions Green-VLA.

During execution, Green-VLA predicts an episode-end probability for the current subtask, when this score exceeds a high threshold (e.g., >0.98), the planner queries a feedback module to evaluate whether the subtask is truly complete. If the VLM judges the subtask as successful, it predicts the next subtask in the sequence; if not, it replans the current subtask and updates the prompt accordingly. Importantly, GigaVision is pretrained and kept frozen in all subsequent Green-VLA training; it is only used at inference time to drive this high-level, language-based task decomposition, feedback, and replanning loop.
\subsection{Unified Action Space for Multi-Embodiment Control}
Most current VLA systems do not use large-scale multi-action space pretraining, or reduce it to naive padding of heterogeneous actions into a single vector. In practice, this often destroys positive transfer. Consider embodiments $e \in \mathcal{E}$ with action spaces
$\mathcal{A}_e \in \mathbb{R}^{d_e}, \quad d_e \text{ varying across robots and control types}$,
and a dataset $\mathcal{D} = \{(x_t^e, a_t^e, e)\}$,
where $x_t^e$ denotes multimodal observations, and $a_t^e \in \mathcal{A}_e$ the corresponding action. A common strategy is to embed all $a_t^e$ into a padded space $\tilde{\mathcal{A}} = \mathbb{R}^{d{\max}}$ via $\tilde{a}_t^e = P_e(a_t^e)$,
where $P_e$ pads actions with zeros up to $d_{\max}$, and then train a single policy $\pi_{\theta}$ with
\begin{equation*} \label{eq:naive-padding}
\mathcal{L}(\theta)
= \mathbb{E}_{(x_t^e, a_t^e)} \big[ \lVert \pi_\theta(x_t^e) - \tilde{a}_t^e \rVert_2^2 \big].
\end{equation*}

This formulation is mis-specified for diverse control spaces:

1.	When two embodiments $e_1, e_2$ use overlapping coordinates differently, the optimal solution to $\mathcal{L}$ averages incompatible targets. Even if $\pi_\theta$ were perfect in predicting $a_t^{e_1}$ in its native control space, it would still be penalized for not matching $\tilde{a}_t^{e_2}$ on the shared (but semantically different) dimensions. This encourages shortcut learning (e.g., inferring "which dataset am I in?" and overfitting to spurious cues) instead of learning embodiment-robust structure.

2.	In many aggregated datasets, the same robot $e$ is logged under multiple control parameterizations (e.g., joint-space torques, joint positions, Cartesian end-effector deltas, etc.), each mapped into $\tilde{\mathcal{A}}$ differently. Under $\mathcal{L}$, the policy is simultaneously trained to match incompatible targets for the same embodiment and state distribution.

Formally, let $m_e \in \{0,1\}^{d_{\max}}$ be the indicator of valid dimensions for embodiment $e$. Then:
\begin{equation*} \label{eq:padding-penalty}
\begin{split}
\mathcal{L}(\theta)
&= \mathbb{E}\Big[
\underbrace{\lVert m_e \odot (\pi_\theta(x_t^e) - \tilde{a}_t^e) \rVert_2^2}_{\text{valid coords}}
 +
\underbrace{\lVert (1 - m_e) \odot (\pi_\theta(x_t^e) - \tilde{a}_t^e) \rVert_2^2}_{\text{spurious penalty}}
\Big],
\end{split}
\end{equation*}
where $\odot$ is Hadamard (element-wise) product. The second term is purely an artifact of padding and directly conflicts with cross-embodiment generalization. Moreover, we may have conflicts in the first valid coordinates across embodiments.

\paragraph{Unified semantic layout.}
Green-VLA replaces naive padding with a \emph{unified action space} $\mathcal{A}_u \subset \mathbb{R}^{64}$ with a fixed semantic layout (right side of \autoref{fig:architecture-full}) where each index range has a consistent physical meaning across all robots that implement it. Here we map native actions (joint, Cartesian, gripper, base, etc.) into the corresponding unified slots with $\Phi_e: \mathcal{A}_e \to \mathcal{A}_u$ and use a binary mask $m_e \in \{0,1\}^{64}$, indicating which slots are used by $e$.

% \begin{figure*}[t] 
%     % floats
%     \centering % Center the image
%     \includegraphics[width=0.8\textwidth]{images/arch/uam.png} % Insert the image file
%     \caption{Unifying heterogeneous robot action spaces into a shared
% 64D action vector $\mathcal{A}_u$. Each robot (Google robot, ALOHA, GR1)
% originally store actions in different layouts (joint / Cartesian / gripper
% / base plus padding). Green-VLA maps semantically corresponding slots
% (hands, base, end-effector pose) into fixed positions in $\mathcal{A}_u$ while masking the remaining dimensions.} % Add a caption
%     \label{fig:unification} % Add a label for referencing
% \end{figure*}

Training then minimizes a \emph{masked} BC objective:
\begin{equation*} \label{eq:unified-bc}
\mathcal{L}_{\mathrm{uni}}(\theta)
= \mathbb{E}
\Big[
\big\| m_e \odot \big( \pi_\theta(x_t^e, c_e) - \Phi_e(a_t^e) \big) \big\|_2^2
\Big],
\end{equation*}

where $c_e$ is an embodiment/control-type prompt. Crucially, no loss is applied to $(1-m_e)$, eliminating spurious gradients from padding and maintaining consistent semantics: for example, the model may predict in joint and Cartesian space and not be penalized for these predictions. This gives our model the ability to learn multi-embodiment semantics, as training in one action space does not destroy semantics for others.

\paragraph{Dynamic embodiment and control-type prompting.}
To make the action mapping explicit and controllable, we condition Green-VLA on a structured control prompt
\begin{equation*} \label{eq:control-prompt}
\begin{split}
c_e = \big(
&\text{\#arms},\; \text{\#hands},\; \text{gripper/dex-hand}, \;\text{joint/cartesian},\; \text{mobile/static},\; \text{slots used}\;
\big),
\end{split}
\end{equation*}
serialized as tokens. The policy thus predicts:
\begin{equation*} \label{eq:policy-predict}
\hat{u}_t = \pi_\theta(x_t^e, c_e) \in \mathcal{A}_u ,
\end{equation*}
and the downstream controller of robot $e$ applies the inverse map $\Phi_e^{-1}$ to the active slots:
\begin{equation*} \label{eq:inverse-map}
\hat{a}_t^e = \Phi_e^{-1}\big(m_e \odot \hat{u}_t\big).
\end{equation*}
This design (i) preserves identifiability of shared skills and (ii) allows control over which action modalities are produced during inference.

To reduce conflicting noise/targets across embodiments on unused slots that introduce unnecessary variance at inference stage for embodiment $e$, we localize the noise injection:
\begin{equation*} \label{eq:noise-localization}
\epsilon_e \sim \mathcal{N}(0, I_{k_e}), \quad
\tilde{\epsilon} = m_e \odot \text{embed}(\epsilon_e).
\end{equation*}

\paragraph{Retargeting for target humanoid.}
Beyond unifying action coordinates, Green-VLA also performs explicit \emph{retargeting} from diverse training robots to a chosen target embodiment, such as a humanoid with dexterous hands. Intuitively, retargeting means, given a source robot (e.g., a 7-DoF arm with a gripper, or a dual-arm platform without fingered hands), we transform its recorded actions into the configuration space of the target humanoid so that “doing the same thing” looks and feels as similar as possible. Concretely, we align semantically corresponding parts-left arm to left arm, right arm to right arm, gripper open/close to the humanoid’s dexterous hand grasp and map them into the appropriate slots of our unified action space. When the degrees of freedom differ, we do not naively pad or duplicate; instead, we choose the closest meaningful match (e.g., mapping a 6-DoF end-effector pose to the nearest feasible joint configuration) so that the target humanoid can imitate the intent of the source trajectory rather than its exact joint layout. This way, demonstrations from heterogeneous anthropomorphic and non-anthropomorphic robots become usable “as-if-humanoid” experience, strengthening transfer and ensuring that every added dataset directly enriches the control repertoire of the deployed system.

\paragraph{Stabilizing training}
To stabilize multi-embodiment training with flow matching, we treat dataset-embodiment sampling as a scheduled mixture: each dataset $i$ has a target weight $w_i$ (from embodiment relevance, size, and \dataqa{} quality), but during training we sample according to:
\begin{equation*} \label{eq:sampling-schedule}
  W_i^{(t)} = \frac{w_i^{\alpha_t}}{\sum_j w_j^{\alpha_t}},
  \qquad
  \alpha_t \in [0,1],\ \alpha_0 = 0,\ \alpha_T = 1,
\end{equation*}

Thus, we start from a uniform mixture ($\alpha=0$) and gradually converge to the desired biased distribution ($\alpha=1$). Intuitively, this prevents early collapse onto a few dominant embodiments and lets the model first learn shared structure before specializing. This curriculum is especially important when using a high-momentum optimizer (e.g., $\beta_1 = 0.95\text{–}0.98$) with large batch size. The effective update is a long-term moving average, so if we begin with a heavily skewed $w_i$, the gradients from large/target datasets quickly dominate the momentum buffer, and rare embodiments are “washed out” even if their $w_i$ are initially large. This leads to mode forgetting and poor cross-robot transfer. Starting balanced ensures all embodiments contribute to the early representation; as $\alpha$ increases, the momentum tracks a smoother, better-conditioned shift toward the target distribution instead of locking onto a single mode.

\begin{figure}[h]
    \centering
    \includegraphics[width=\linewidth]{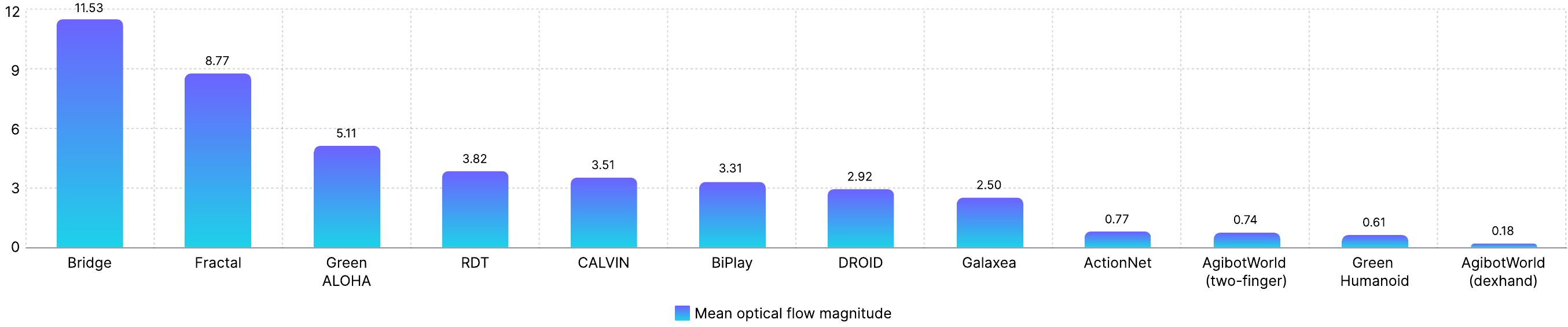}
    \caption{Mean optical flow magnitude per dataset used for temporal alignment (higher values correspond to faster apparent motion).}
    \label{fig:optical-flow}
\end{figure}

\paragraph{Action alignment}
To further standardize heterogeneous demonstrations, Green-VLA performs action alignment to equalize the varying execution speeds across datasets. Different robots and operators move at different intrinsic speeds—some execute trajectories quickly and smoothly, while others move slowly due to task complexity or teleoperation system properties. Mixing such datasets naively confuses the model about the expected magnitude of dynamics. We address this issue by resampling trajectories and interpolating actions using monotonic cubic splines to normalize the effective motion per step. To determine the appropriate resampling rate for each dataset, we estimate its execution speed by computing the mean optical flow magnitude from wrist cameras as a proxy for visual motion (See \autoref{fig:optical-flow}). For example, datasets with low capture frequency, like Bridge and Fractal, tend to have large between-frame optical flow, so we densify their actions with additional interpolated waypoints. On the other hand, we speed up datasets with high capture frequency and slow motion like AgiBot DexHand by downsampling actions. This flow-based resampling creates temporally aligned trajectories in the unified action space, where similar visual and geometric changes correspond to similar action increments, making cross-dataset generalization more robust. In practice, we apply this alignment both to open-source datasets and to our own teleoperation data to ensure consistent motion statistics across all sources.

% \begin{table}[h!]
% \centering
% \small
% \setlength{\tabcolsep}{4pt}
% \caption{Mean optical flow magnitude per dataset used for temporal alignment
% (higher values correspond to faster apparent motion).}
% \label{tab:optical-flow}
% \begin{tabular}{@{}lc@{}}
% \toprule
% \textbf{Dataset} & \textbf{Mean flow} \\ % arbitrary units
% \midrule
% AgibotWorld (dexhand)      & 0.18 \\
% AgibotWorld (two-finger)   & 0.74 \\
% Aloha                      & 5.11 \\
% Aloha-ecom                 & 3.44 \\
% BiPlay                     & 0.83 \\
% CALVIN                     & 3.51 \\
% Centaur (dexhand)          & 0.55 \\
% DROID                      & 2.92 \\
% Humanoid (Green)           & 0.61 \\
% RDT                        & 3.82 \\
% Galaxea                    & 2.50 \\
% Bridge                     & 11.53 \\
% Fractal                    & 8.77 \\
% ActionNet                  & 0.77 \\
% \bottomrule
% \end{tabular}
% \end{table}
% TODO: после правок дизайнера проверить список датасетов. Например, aloha-ecom не должно быть. Название Mean Flow мб не лучшее (обозначить скорость как-то в тексте или написать Mean Optical Flow Magnitude, например).
% BiPlay поправить, это у нас видосы замедленные в 4 раза были, лучше вместо 0.83 напиметь 3.32, получится близко к другим алохам.

\paragraph{Temporal Scale Conditioning for Long–Short Horizon Control}
On top of temporally aligned trajectories, we introduce a speed-conditioned augmentation that lets Green-VLA represent both fine-grained manipulation and faster, coarse motions within the same model. For each training sample, we draw a scalar “action speed” factor $v \sim p(v)$ and use our interpolation / resampling procedure to warp the target trajectory: values $v > 1$ effectively slow the sequence down (more intermediate waypoints), while $v < 1$ compress it (fewer, coarser steps). This changes the effective temporal resolution of the supervision and augments the target distribution without breaking cross-dataset consistency.

Concretely, let $h_t$ be the hidden state of the action expert. We apply a $v$-conditioned RMS-style modulation:
\begin{equation*} \label{eq:rms-modulation}
\tilde{h}_t = \mathrm{RMSNorm}(h_t), \quad
\hat{h}_t = \gamma(v)\,\tilde{h}_t + \beta(v),
\end{equation*}
and predict actions from $\hat{h}_t$, where $\gamma(v), \beta(v)$ are small learned functions of $v$. In this way, each sample in the batch is trained at a different effective speed, and the model explicitly knows which regime it should operate in. Because all datasets have been normalized to a common motion scale using optical-flow–based alignment, the meaning of $v$ is consistent across embodiments: higher values emphasize slow, high-resolution adjustments, while lower values encourage faster progress along the trajectory. At inference time, $v$ becomes a simple hyperparameter that trades off local precision versus long-horizon efficiency, and in future work it can be set automatically by a high-level robotics planner.

This speed-conditioned modulation effectively teaches Green-VLA to operate on multiple temporal “zoom levels,” which is exactly what we want for balancing short- and long-horizon capabilities. When $v$ is low, the model is encouraged to make small, incremental changes between steps—capturing detailed contact dynamics. When $v$ is large, the same underlying representation is pushed to cover larger progress per step, learning stable shortcuts through easy segments like reaching, lifting, or base motion.  At inference time, this translates into a controllable continuum: the same Green-VLA can act as a careful local controller in delicate phases and as an efficient high-level executor over long horizons, without retraining or separate planners. Now Green-VLA is controllable such that $v$ is a hyperparameter of inference, however, as future work, a high-level robotics VLA may control this parameter.

\begin{figure*}[ht!]
    \centering
    \includegraphics[width=0.95\textwidth]{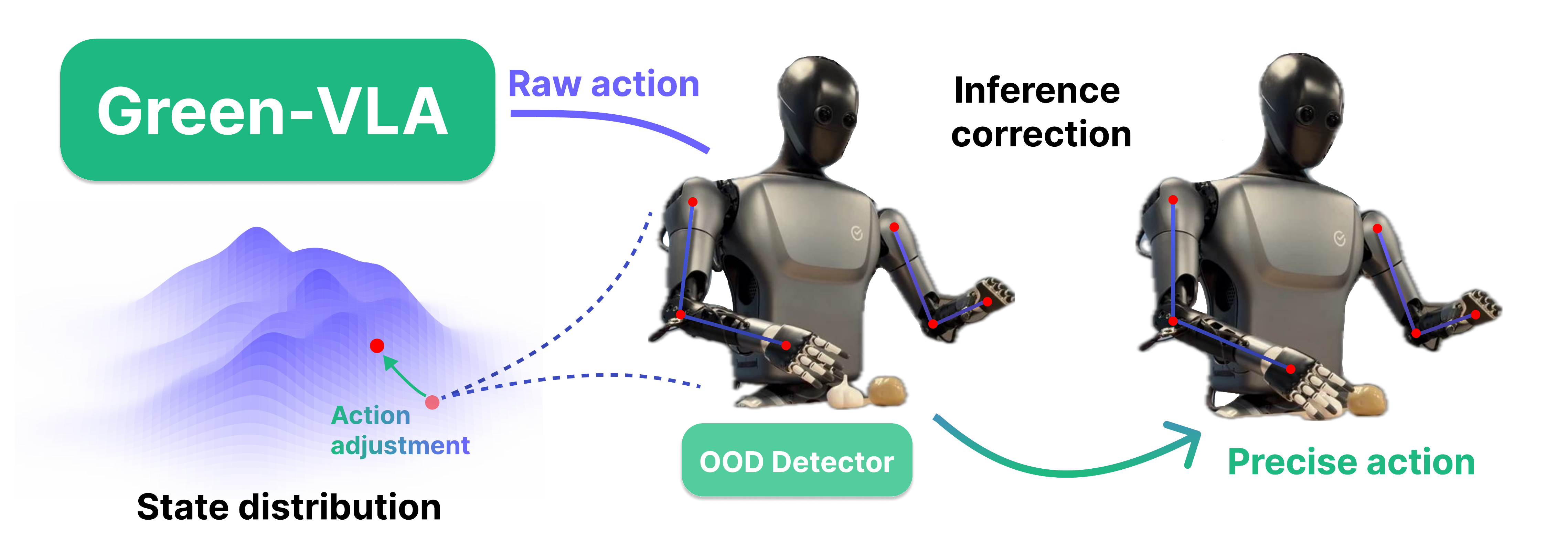}
    \caption{Robot state density is modeled with a Gaussian Mixture Model (GMM). Actions that would cause the robot to enter an out-of-distribution state with low GMM density are corrected based on the GMM density gradient.}
    \label{fig:ood}
\end{figure*}

\paragraph{Episode progress}
The flow-matching action expert jointly predicts a normalized episode progress signal $\hat{\rho}_t \in [0,1]$, trained from demonstrations with target $\rho_t = t / T$, where $t$ is the current step and $T$ is the episode length. This scalar is exposed to downstream planners, enabling them to request updated instructions or subgoals as long-horizon tasks unfold.

\paragraph{OOD detector}
Inspired by \cite{back_to_the_manifold}, we further equip the architecture with an online out-of-distribution (OOD) detector over the predicted trajectory, modeled as a mixture of Gaussians fitted on robot states from the Green-VLA training set:
\begin{equation*} \label{eq:ood-detector}
p_{\text{train}}(s) = \sum_{k} \phi_k \,\mathcal{N}(s \mid \mu_k, \Sigma_k).
\end{equation*}
At inference, if the predicted actions cause the robot to enter a state $s$ such that $p_{\text{train}}(s) < \tau_{\text{ood}}$, we recompute the actions to instead reach a corrected state $s + \alpha\nabla p_{\text{train}}(s)$, where $\alpha = 0.2$ is the gradient step size (see \autoref{fig:ood}). Overall, this module nudges the policy back toward the training-state distribution, improving its safety and stability in long tasks.

\subsection{Guidance with joint prediction module}
In dynamic settings such as e-commerce shelves, new or rarely seen items may appear that are not recognized by VLA, yet are explicitly specified in the instruction (e.g., “pick the blue 500ml bottle of X”). To handle this, Green-VLA augments its policy with a lightweight training-free guidance module. First, a joint prediction head takes the current observation and language instruction and predicts a target point $p^\star$ in the robot’s workspace corresponding to the described item. Then, during action generation, we apply flow-matching guidance: instead of sampling from the unconditional velocity field $v_\theta(x)$, we bias the field toward trajectories that move the end-effector toward $p^\star$

\paragraph{Joint prediction module}
To estimate the target point $p^\star$ required by the guidance mechanism, we propose the Joint Prediction Module (JPM). The key idea is to decompose the manipulation instruction into subtasks and infer a goal robot configuration directly from each subtask’s textual description.

Given an observation image and an instruction specifying which object should be manipulated, JPM first queries a specialized Vision-Language Model to predict a 2D affordance point $(u, v)$ on the image via a pointing-based mechanism. This point corresponds to the location most relevant for grasping or interaction.

We then lift this 2D affordance into the robot's 3D workspace. For a pixel $(u, v)$ with depth $d(u,v)$, camera intrinsic matrix $K$, and camera pose $T_{c}^{w} \!\in\! SE(3)$, the corresponding 3D point in the world (robot-base) frame is
\begin{equation*} \label{eq:2d-to-3d}
\begin{bmatrix} {p}^\star \\ 1 \end{bmatrix}
=
T_c^w
\begin{bmatrix}
d(u,v)\,K^{-1}[u,\,v,\,1]^\top \\ 1
\end{bmatrix}.
\end{equation*}

% \begin{figure*}[t!]
%     \centering
%     \includegraphics[width=0.7\textwidth]{images/R2/DSRL.drawio.png}
%     \caption{Phase R2: Optimization of source distribution}
%     \label{fig:dsrl}
% \end{figure*}

% \begin{figure*}[t!]
%     \centering
%     \includegraphics[width=0.7\textwidth]{images/R2/arch_parl.png}
%     \caption{TODO? check if needed}
%     \label{fig:parl}
% \end{figure*}

The resulting point $p^\star \in \mathbb{R}^3$ represents the desired target location in the robot’s workspace and is fully consistent with the scene geometry and the language query.

Finally, JPM computes a feasible joint configuration $q^\star$ that positions the end-effector at $p^\star$ by solving an inverse kinematics (IK) problem with typical constraints on orientation, reachability, and collision avoidance. The resulting $q^\star$ is then injected into the policy and used during flow-matching guidance to bias the velocity field $v_\theta(x)$ toward trajectories that move the end-effector toward $p^\star$.

\paragraph{Guidance module}
For conditioning, we adopt a pseudoinverse guidance scheme ($\Pi\mathrm{GDM}$; \cite{Song23}) on top of the learned flow-matching policy. At each denoising step, we modify the velocity field $v$ with an additional guidance term that nudges the trajectory toward a desired target value $Y$ (e.g., the predicted grasp point or object-specific action). Concretely, $\Pi\mathrm{GDM}$ estimates how changes in the current noisy action $\mathbf{A}_t^\tau$ affect the final clean action and injects a gradient-based correction in that direction, yielding a guided field $\mathbf{v}_{\Pi\mathrm{GDM}}$ that preserves the prior while steering the generation so that the final output is consistent with the inferred target constraint.

\subsection{RL fine-tuning (R2)}
A model that has completed the full L0–L1–R0–R1 training cycle is further fine-tuned using reinforcement learning. We employ two approaches: Trajectory optimization with native fine-tuning inspired by \cite{mark24}, and Optimization of source distribution \cite{wagenmaker25}.

\paragraph{Trajectory optimization with native fine-tuning} A small separate critic estimating state–action pair value (the Q-function) is trained on the R1 dataset with sparse rewards. We use Implicit Q-Learning \cite{kostrikov21} for its stability, elegant mitigation of the Q-value overestimation problem, and smooth offline-to-online transition. The critic is trained according to the following objectives:
\begin{equation*} \label{eq:iql-value}
L_V(\psi) = \mathbb{E}_{(s,a)\sim D_{\text{off}}}
\big[L_2^{\tau}(Q_{\hat{\theta}}(s,a) - V_\psi(s))\big],
\end{equation*}
\begin{equation*} \label{eq:iql-q}
L_Q(\theta) = \mathbb{E}_{(s,a,s')\sim D_{\text{off}}}
\big[(r(s,a) + \gamma V_\psi(s') - Q_\theta(s,a))^2\big].
\end{equation*}
The parameter $\tau\in(0,1)$ is the expectile of some random variable $X$ defined as a solution to the asymmetric least squares problem
\begin{equation*} \label{eq:expectile}
\arg\min_{m_\tau}\mathbb{E}_{x \sim X}[L_2^\tau(x-m_\tau)],
\end{equation*}
where $L_2^\tau(u)=\lvert \tau-\mathbf{1}(u<0)\rvert u^2$.

Then, the base model that has undergone R1 fine-tuning generates trajectories in the environment—this may be either a simulator environment such as CALVIN \cite{calvin} or Simpler, or a real-robot setup.
The recorded trajectories are improved using the trained Q-function: at each step of the trajectory, the gradient of the Q-function with respect to the action, $\nabla_a Q(s,a)$, is computed.
This gradient is normalized and added to the original action generated by the base model at the current step, scaled by a multiplier chosen as a hyperparameter (which, in a certain sense, is analogous to a learning rate):
\begin{equation*} \label{eq:qgrad-update}
a \leftarrow a 
+ \eta \, 
\frac{\nabla_a Q(s,a)}{\|\nabla_a Q(s,a)\|},
\end{equation*}
The gradient computation and update are performed N times (another hyperparameter), resulting in an optimized trajectory.

Since the trained Q-function is not optimal, and the hyperparameters are tuned heuristically, the resulting action is not guaranteed to improve the trajectory (we consider an improvement to be a reduction in task execution time) and may even worsen the outcome.
To prevent adding low-quality data to the dataset, we perform validation of the optimized trajectories in the environment: the operator restores the environment to the state corresponding to the beginning of the original trajectory rollout, after which the improved trajectory is executed and the result is saved.
The enriched data are added to the R1 dataset, after which native R1 fine-tuning is repeated starting from the weights obtained at the end of phase R0.

The advantage of trajectory optimization with native fine-tuning is that we do not need to modify the weights of the base model using gradients obtained from executing any RL algorithm directly.
Because the base models we use are primarily flow-matching models, traditional RL fine-tuning would face several challenges. First, on-policy PG methods like PPO or GRPO require estimating the log-probabilities of generated actions. There are different approaches to handle this, however, all of them have some limitations affecting either the wall-clock training time, training stability, or representational capability of a model. Off-policy methods are not an easy alternative either, as they require backpropagating value gradients through the iterative action generation process which may be unstable and demand meticulous tuning.

\begin{figure*}[t!]
    \centering
    \includegraphics[width=0.9\textwidth]{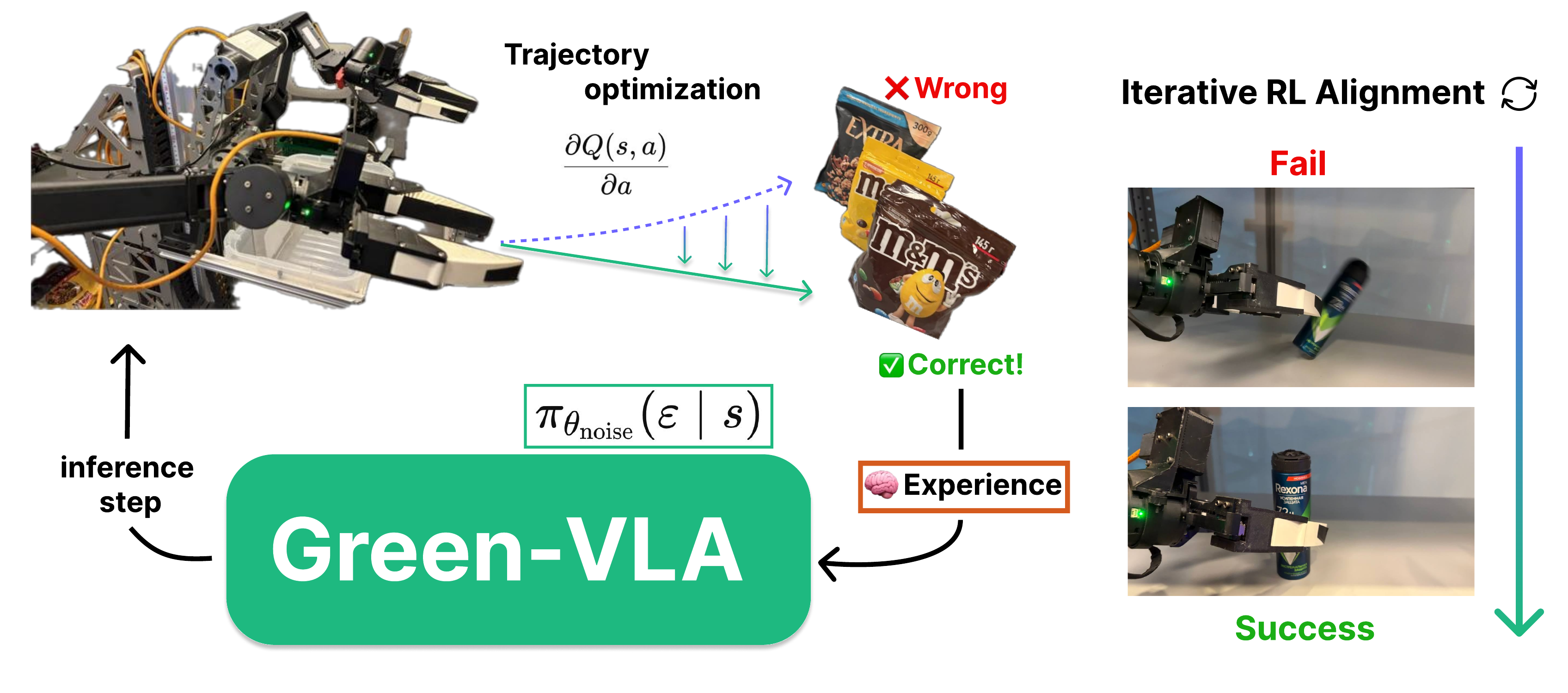}
    \caption{Phase R2: RL alignment. Optimization of the source noise distribution: an actor $\pi_{\theta_{noise}}(\epsilon | s)$ learns to sample noise that improves the flow-matching policy’s actions. PARL-style trajectory optimization: experience and teleoperation data are iteratively refined with Q-function gradients and added to train set.}
    \label{fig:R2fig}
\end{figure*}

\paragraph{Optimization of source distribution} After improving model performance with optimized trajectories, we apply the second RL fine-tuning approach, which also has the advantage of preserving the base model weights from being altered directly.
Since a flow-matching model generates actions based on vectors sampled from some source distribution, e.g. Gaussian, the generation result depends on the initialization of these vectors — or, more formally, on the distribution from which they are sampled. The goal of flow-matching is training a velocity field neural network with some set of parameters $\theta$ that determines a time-dependent flow, defined as 
\begin{equation*} \label{eq:flow-matching}
\frac{\mathrm{d}}{\mathrm{d}t}\psi_t(x)=u_\theta(\psi_t(x),t),
\end{equation*}
where $\psi_t(x) := \psi(t, x)$ and $\psi_0(x) = x$. Given this vector field, samples from the target distribution $X_1 = \psi_1(X_0)$, with $X_0 \sim p_0$, can be obtained by solving the corresponding ODE \cite{lipman24}.
The R1 phase of training uses isotropic Gaussian distribution as $p_0$ - the source of random vectors; accordingly, our base model learns a transformation from this distribution to the action distribution in the R1 dataset.
Using an actor–critic algorithm, we train a small separate actor network that generates the random noise fed into the base model, which maximizes the return obtained by the base model in the environment. In other words, the actor provides a new distribution $p_0'$ which purpose is improving the base model behavior. This is an online algorithm requiring RL-guided trajectory execution in the environment. However, compared to traditional online RL algorithms, it is safer in terms of controlled exploration: the actions generated by the base model tend to remain close to those in the training dataset, even if the distribution of random noise shifts.

\section{Experiment Metrics Across Phases}
We evaluate Green-VLA across staged training phases (R0, R1, R2) and heterogeneous embodiments, focusing on task-following, long-horizon robustness, and efficiency. Overall, Green-VLA is a mid-scale $\sim$5B-parameter VLA model in its latest configuration. We use \textbf{Qwen3-VL-4B-Instruct} as the primary vision--language backbone due to its higher VLM quality, and couple it with a dedicated flow-matching action expert and lightweight heads that account for the remaining parameters. In earlier versions we used PaliGemma (3B), resulting in an overall $\sim$4B-parameter VLA. Our architecture cleanly separates semantic perception from high-frequency control by attaching the action expert on top of the frozen or lightly tuned VLM trunk. For the R0 robotics pretraining stage, we train Green-VLA for over $10^5$ optimization steps on a cluster of 64 H100 GPUs, which is sufficient to fully saturate the unified multi-embodiment robotic dataset while remaining substantially lighter in both data and compute than prior very large VLA systems.

First, in the \textbf{R0 phase}, we benchmark table-cleaning, pick-and-place tasks on the AgileX Magic Cobot \cite{cobotmagic_agilex}, measuring success rate (SR) and execution efficiency (time-to-clear and actions-to-clear) against $\pi_0$, GR00T N1, AgiBot GO-1, and WALL-OSS. Importantly, Green-VLA is trained on $\sim3{,}000$ hours of unified demonstrations—substantially less than the $>10{,}000$ hours of data used in $\pi_0$—allowing us to quantify the benefits of quality alignment and unified actions under constrained data. Second, we report R0 performance on standardized open benchmarks derived from WidowX  and Google-robot, comparing SR against Flower, RT-1X, $\pi_0$, OpenVLA and the Dexbotic \cite{Dexbotic} version of MemoryVLA \cite{memvla} to assess cross-embodiment generalization under identical protocols. Third, in the \textbf{R1 phase}, we fine-tune Green-VLA on the CALVIN environment using only our generic R0-pretrained checkpoint (i.e., without ever pretraining on CALVIN data), and measure multitask SR and compositional generalization isolating the effect of embodiment-specific adaptation.

\begin{figure*}[t!]
    \centering
    \begin{subfigure}[t]{0.3\textwidth}
        \centering
        \includegraphics[width=\textwidth]{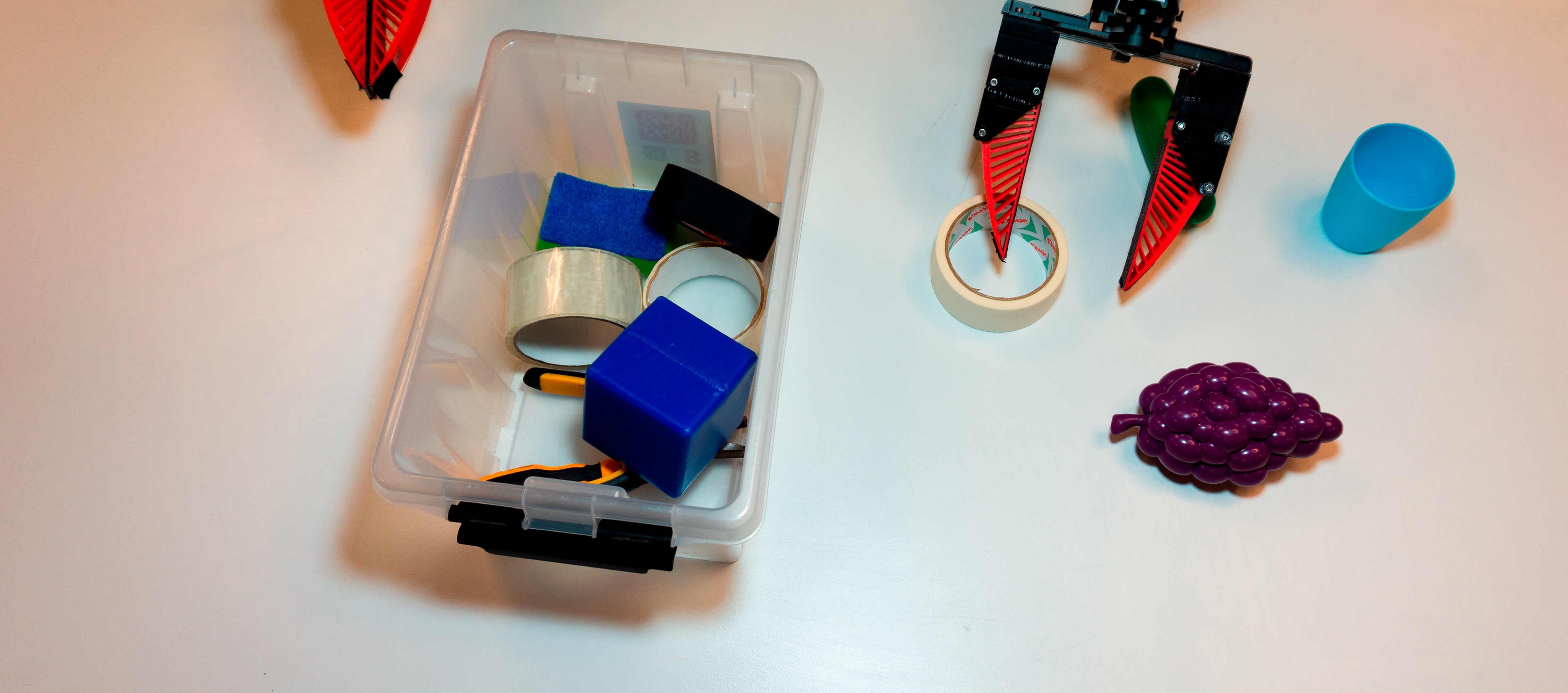}
        \caption{Setup for "pick tape" task.}
        \label{fig:sphere}
    \end{subfigure}
    \hfill
    \centering
    \begin{subfigure}[t]{0.3\textwidth}
        \centering
        \includegraphics[width=\textwidth]{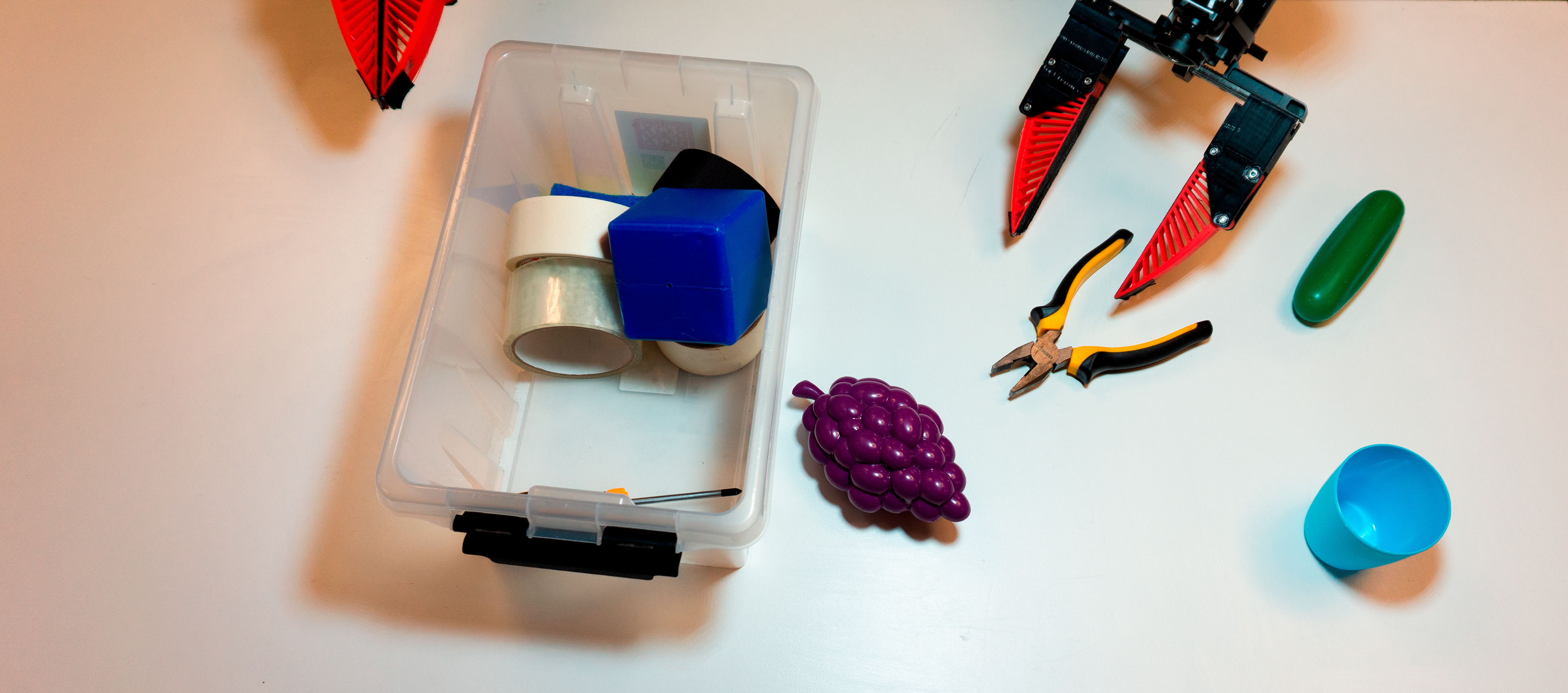}
        \caption{Setup for "pick pliers" task.}
        \label{fig:sphere}
    \end{subfigure}
    \hfill
    \centering
    \begin{subfigure}[t]{0.3\textwidth}
        \centering
        \includegraphics[width=\textwidth]{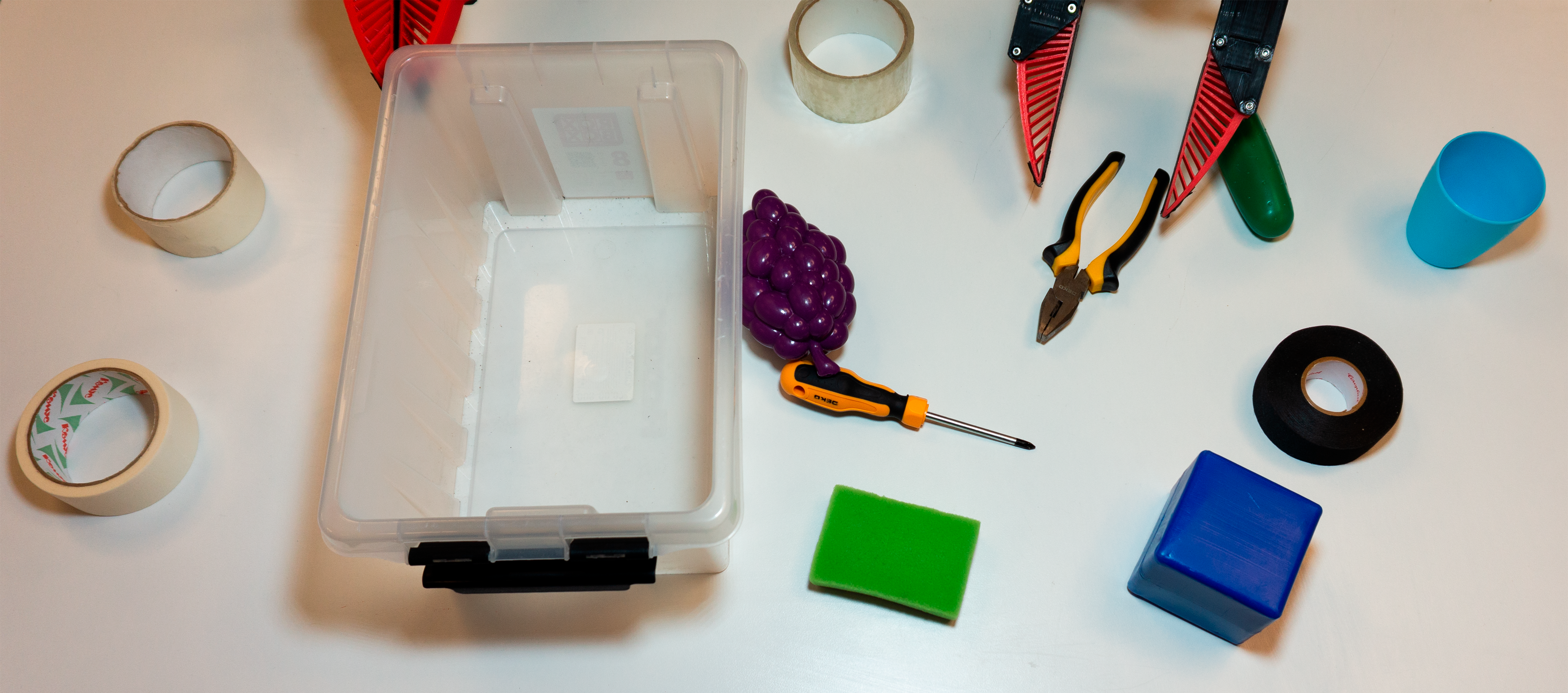}
        \caption{Cleaning table setup}
        \label{fig:sphere}
    \end{subfigure}
    \caption{a), b) Examples of task-following setup. c) Example of full table cleaning setup}
    \label{fig:benchmark_grid}
\end{figure*}

\paragraph{Bimanual cleaning table setup.}
For manipulation on Cobot Magic, we evaluate instruction-conditioned picking and table-cleaning in a controlled, reproducible environment shared across $\pi_0$, WALL-OSS, AgiBot GO-1, GR00T N1, and Green-VLA from R0 phase only, without any fine-tuning. We additionally fine-tune all the models used as baselines for 20,000 iterations on a dataset containing ALOHA trajectories for the table-cleaning task. In the single-target setting, we sample 4 to 6 objects on the table and provide a natural-language instruction of the form "Pick the \texttt{[specified item]} and place it in the box." For each candidate object, we run 10 episodes with randomized placements and distractors (identical across methods), see \autoref{fig:benchmark_grid}, and report the fraction of trials where the correct item is grasped and successfully dropped into the box \autoref{tab:stat-bench}. In the full-task setting, we place 15 to 20 objects on the table and issue sequential instructions for multiple items and run the evaluation 10 times. We measure (i) first-attempt correctness—the fraction of instructions where the first pick matches the requested item—and (ii) the mean time (or action steps) required to completely clear all instructed items into the box. This protocol jointly captures task following, object disambiguation under clutter, and execution efficiency, enabling a direct comparison of our R0-stage Green-VLA against prior VLA baselines. We can see that Green-VLA achieves higher task-following accuracy and SR than $\pi_0$. Unexpectedly, we find that multi-embodiment pretraining yields strong results for almost all datasets in the training set, and the model does not require additional tuning for a specific embodiment on the same data.

\begin{table*}[t!]
\centering
\small
\setlength{\tabcolsep}{4pt}
\caption{ALOHA table-cleaning task}
\label{tab:stat-bench}
\begin{tabular}{@{}l
                S[table-format=2.1]
                S[table-format=2.1]
                S[table-format=2.1]
                S[table-format=2.1]
                l@{}} % <- left align AVG Time
\toprule
\multicolumn{1}{l}{\textbf{Policy}} &
\multicolumn{1}{c}{\textbf{Tape}} &
\multicolumn{1}{c}{\textbf{Screwdrivers}} &
\multicolumn{1}{c}{\textbf{Pliers}} &
\multicolumn{1}{c}{\textbf{First item SR}} &
\multicolumn{1}{c}{\textbf{AVG Time}} \\
\midrule
$\pi_0$      & \cellcolor{top3color}46.3 & 29.7 & \cellcolor{top3color}31.8 & \cellcolor{top3color}35.6 & \cellcolor{top2color}2m59s\\
GR00T N1    & 38.9 & \cellcolor{top3color}35.4 & 29.5 & 33.2 & >5m \\
WALL-OSS     & 27.4 & 14.2 & 27.3 & 12.1 & >5m \\
AgiBot GO-1  & \cellcolor{top2color}57.8 & \cellcolor{top2color}48.6 & \cellcolor{top2color}33.2 & \cellcolor{top2color}38.4 & \cellcolor{top3color}3m57s \\
Green-VLA(R0)        & \cellcolor{top1color}83.1 & \cellcolor{top1color}52.1 & \cellcolor{top1color}63.7 & \cellcolor{top1color}69.5 & \cellcolor{top1color}1m35s \\
\bottomrule
\end{tabular}
\end{table*}

\paragraph{Simpler.}
For the Simpler benchmarks on WidowX and Google Robot, we compare our R0-pretrained Green-VLA against a range of existing foundation policies, including $\pi_0$ (R0 and SFT-fine-tuned), OpenVLA \cite{openvla}, Flower \cite{flower2025}, RT-1X \cite{rt1x_openxembodiment} (on RT-1 data), X-VLA \cite{xvla}, Magma \cite{magma}, GR00T-N1.6 \cite{groot_n16}, DB-MemVLA \cite{db_memvla} and EO-1 \cite{eo1}. Our R0 model outperforms other methods at the same pure-pretrain stage and achieves performance comparable to several fine-tuned baselines. We run all experiments with the default Simpler horizon (80-200 steps for Google Robot and 80 steps for WidowX) and use our episode-end prediction head (Green-VLA EEP) to terminate execution once the task is judged complete. This is particularly important for Google Robot tasks: unnecessary motion after the goal is reached can easily turn a successful configuration into a failed one, mirroring real-world deployments where the robot should complete the task and then wait for a new instruction instead of “fidgeting” in the scene. For Simpler, we report the mean result aggregated across 7 evaluation.

\begin{table*}[ht!]
\centering
\scriptsize
\setlength{\tabcolsep}{2pt}
\caption{SimplerEnv evaluation across different policies on Google Robot tasks for default number of Simpler episode steps. We report models that were pretrained on a mixture-data including Rt-1 dataset or fine-tuned on it. }
\label{tab:fractal-bridge}
\begin{tabular*}{\textwidth}{@{\extracolsep{\fill}}lccccccccccc@{}} % 1 + 11 = 12 columns
\toprule
& \multicolumn{5}{c}{\textbf{Visual Matching}}
& \multicolumn{5}{c}{\textbf{Variant Aggregation}}
& \multicolumn{1}{c}{\textbf{\#Overall}} \\
\cmidrule(lr){2-6}\cmidrule(lr){7-11}\cmidrule(lr){12-12}
\textbf{Model} &
\textbf{Drawer} & \textbf{Move near} & \textbf{Pick Coke} & \textbf{Apple} & \textbf{AVG VM} &
\textbf{Drawer} & \textbf{Move near} & \textbf{Pick Coke} & \textbf{Apple} & \textbf{AVG VA} &
\textbf{Average} \\
\midrule
$\pi_0$ (Fine-tune) & 38.3 & 65.3 & 72.7 & 0.0 & 44.1 & 25.6 & 63.7 & 75.2 & 0.0 & 41.1 & 42.6 \\
$\pi_{0.5}$ (Fine-tune) & 57.9 & 72.5 & 86.7 & 0.0 & 54.3 & 50.5 & 73.5 & 87.4 & 0.0 & 52.8 & 53.6 \\
X-VLA & \cellcolor{top3color}64.4 & \cellcolor{top1color}84.6 & 93.7 & 18.5 & \cellcolor{top3color}65.3 & 43.7 & \cellcolor{top2color}78.8 & \cellcolor{top2color}96.1 & 30.7 & \cellcolor{top3color}62.3 & \cellcolor{top3color}63.8 \\
GR00T-N1.6 & 61.1 & 73.8 & \cellcolor{top3color}95.3 & 13.0 & 60.8 & \cellcolor{top3color}59.5 & 68.3 & 89.6 & 23.3 & 60.2 & 60.5 \\
Magma & 62.5 & 68.3 & 74.3 & 13.0 & 54.5 & \cellcolor{top2color}60.3 & \cellcolor{top3color}76.9 & 70.8 & \cellcolor{top3color}32.8 & 60.2 & 57.4 \\
EO-1 & \cellcolor{top1color}71.3 & \cellcolor{top2color}83.8 & \cellcolor{top2color}98.0 & \cellcolor{top2color}52.8 & \cellcolor{top2color}76.5 & \cellcolor{top1color}91.6 & \cellcolor{top1color}81.7 & 55.0 & 23.8 & \cellcolor{top2color}63.0 & \cellcolor{top2color}69.8 \\
OpenVLA & 35.6 & 46.2 & 16.3 & 0.0 & 24.5 & 17.7 & 47.7 & 54.5 & 0.0 & 30.0 & 27.2 \\
RT-1-X & 59.7 & 31.7 & 56.7 & \cellcolor{top3color}40.7 & 47.2 & 49.0 & 32.3 & 29.7 & \cellcolor{top2color}40.7 & 37.9 & 42.6 \\
Flower & - & - & - & - & - & -- & -- & -- & -- & -- & 42.4 \\
\midrule
\multicolumn{12}{@{}l}{\textbf{Green-VLA} \hspace{3pt} {\scriptsize (\textit{Paligemma 3B})}} \\[1pt]
Green-VLA(R0) & 62.9 & 61.2 & 90.4 & 0.0 & 53.6 & 33.5 & 38.1 & 75.5 & 0.0 & 36.7 & 45.1 \\
Green-VLA(R1) & 47.0 & 58.7 & 95.0 & 0.0 & 50.1 & 34.1 & 42.9 & 92.1 & 16.0 & 46.2 & 48.1 \\
Green-VLA(R2) & 61.0 & 50.8 & \cellcolor{top1color}98.1 & 0.0 & 52.4 & 51.6 & 71.2 & \cellcolor{top1color}98.2 & 28.0 & \cellcolor{top3color}62.3 & 57.3 \\
\midrule
\multicolumn{12}{@{}l}{\textbf{Green-VLA} \hspace{3pt} {\scriptsize (\textit{Qwen3-VL-4B-Instruct})}} \\[1pt]
Green-VLA(R1) & \cellcolor{top2color}64.8 & \cellcolor{top3color}75.8 & 85.7 & \cellcolor{top1color}81.5 & \cellcolor{top1color}77.0 & 35.7 & 71.9 & \cellcolor{top3color}92.6 & \cellcolor{top1color}66.7 & \cellcolor{top1color}66.7 & \cellcolor{top1color}71.8 \\
\bottomrule
\end{tabular*}
\end{table*}

On WidowX, we perform a three-stage comparison using the same architecture: (i) R0 pretraining on the unfiltered BRIDGE mixture, (ii) R1-style SFT on a filtered BRIDGE subset, and (iii) R2 RL fine-tuning starting from the filtered R1 checkpoint. We observe a consistent increase in success rate across these stages, illustrating how dataset curation plus RL alignment builds on the unified pretrain to progressively improve performance on a concrete target embodiment.

For Flower, OpenVLA, and RT-1X, we report the officially published Simpler results from the FLOWER paper~\cite{flower2025} and the OpenVLA report~\cite{ddvla}, rather than re-implementing these baselines. 

Another models we evaluate on WidowX tasks under the same conditions as Green-VLA for 7 runs and aggregate results using default episode lengths of 80 steps.

\begin{table*}[ht!]
\centering
\scriptsize
\setlength{\tabcolsep}{3pt}
\caption{SimplerEnv evaluation across different policies on WidowX Robot tasks. We report the results for a model that was pretrained on a mixture-data including the Bridge dataset.}
\label{tab:fractal-bridge-widowx}
\begin{tabular*}{\textwidth}{@{\extracolsep{\fill}}lcccccccccc@{}}
\toprule
& \multicolumn{5}{c}{\textbf{Pick}}
& \multicolumn{5}{c}{\textbf{Success}} \\
\cmidrule(lr){2-6}\cmidrule(lr){7-11}
\textbf{Model} &
\textbf{Spoon} & \textbf{Cubes} & \textbf{Eggplant} & \textbf{Carrot} & \textbf{AVG Grasp} &
\textbf{Spoon} & \textbf{Cubes} & \textbf{Eggplant} & \textbf{Carrot} & \textbf{AVG Success}\\
\midrule
$\pi_{0}$ (Fine-tune) & 45.8 & 50.0 & 91.6 & 25.0 & 53.1 & 29.1 & 16.7 & 62.5 & 0.0 & 27.1 \\
$\pi_{0.5}$ (Fine-tune) & 66.7 & 16.7 & 50.0 & 50.0 & 45.9 & 29.2 & 0.0 & 41.7 & 41.7 & 28.2 \\
OpenVLA & 4.1 & 12.5 & 8.3 & 33.0 & 14.5 & 0.0 & 0.0 & 4.1 & 0.0 & 1.0 \\
RT-1-X & 16.7 & 8.3 & 0.0 & 20.8 & 11.5 & 0.0 & 0.0 & 0.0 & 4.2 & 1.1 \\
Flower & -- & -- & -- & -- & -- & 71.0 & 8.0 & 88.0 & 13.0 & 45.0 \\
DB-MemVLA & \cellcolor{top2color}91.7 & 83.3 & 79.2 & \cellcolor{top1color}100.0 & 88.6 & \cellcolor{top3color}85.1 & 57.6 & \cellcolor{top1color}100.0 & 50.0 & \cellcolor{top3color}73.2 \\
X-VLA & \cellcolor{top1color}95.8 & 79.2 & 62.5 & 75.0 & 78.1 & \cellcolor{top1color}91.7 & 37.5 & 62.5 & \cellcolor{top3color}70.8 & 65.6 \\
Magma & 70.8 & 75.0 & \cellcolor{top3color}91.7 & 37.5 & 68.8 & 54.2 & 29.2 & 83.3 & 33.3 & 50.0 \\
GR00T-N1.6 & 58.3 & 20.8 & \cellcolor{top1color}100.0 & 54.2 & 58.3 & 41.7 & 0.0 & 62.5 & 33.3 & 34.4 \\
EO-1 & - & - & - & - & - & 63.6 & \cellcolor{top1color}81.8 & 90.9 & 54.5 & 72.7 \\

\midrule % separator between non-GreenVLA models and GreenVLA block
\multicolumn{11}{@{}l}{\textbf{Green-VLA} \hspace{3pt} {\scriptsize (\textit{Paligemma 3B})}} \\[1pt]
Green-VLA (R0) & 66.7 & \cellcolor{top3color}91.7 & \cellcolor{top3color}91.7 & 50.0 & 75.0 & 33.3 & 33.3 & 88.5 & 25.0 & 45.0 \\
Green-VLA (R1) & 75.0 & \cellcolor{top3color}91.7 & 87.5 & 50.0 & 76.1 & 66.7 & 37.5 & 79.2 & 37.5 & 55.2 \\
Green-VLA (R2) & 87.5 & \cellcolor{top2color}95.8 & \cellcolor{top3color}91.7 & \cellcolor{top2color}91.6 & \cellcolor{top2color}91.7 & \cellcolor{top2color}90.1 & 52.6 & 84.8 & \cellcolor{top1color}89.0 & \cellcolor{top2color}79.1 \\

\midrule
\multicolumn{11}{@{}l}{\textbf{Green-VLA} \hspace{3pt} {\scriptsize (\textit{Qwen3-VL-4B-Instruct})}} \\[1pt]
Green-VLA (R1) & \cellcolor{top2color}91.7 & \cellcolor{top3color}91.7 & \cellcolor{top1color}100.0 & 75.0 & \cellcolor{top3color}89.6 & 79.2 & \cellcolor{top3color}58.3 & \cellcolor{top3color}91.7 & 62.5 & 72.9 \\
Green-VLA (R2) & \cellcolor{top3color}90.6 & \cellcolor{top1color}99.0 & \cellcolor{top2color}99.0 & \cellcolor{top3color}89.6 & \cellcolor{top1color}94.6 & 80.2 & \cellcolor{top2color}70.8 & \cellcolor{top2color}94.8 & \cellcolor{top2color}76.1 & \cellcolor{top1color}80.5 \\
\bottomrule
\end{tabular*}
\end{table*}

\subsection{R1: Efficient Fine-Tuning}

\begin{figure}[H]
    \centering
    \includegraphics[width=\linewidth]{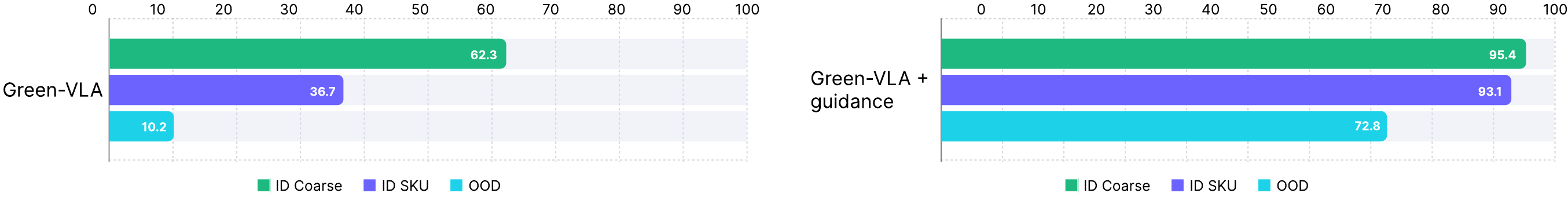}
    \caption{E-commerce shelf picking: top-1 success rate (\%) for Green-VLA with and without JPM guidance. Columns report In-Domain—Coarse (brand/category), In-Domain—SKU (exact variant), and Out-of-Domain (unseen SKUs/packaging). Higher is better.}
    \label{fig:jpm_res}
\end{figure}

\paragraph{Guidance with JPM}
To comprehensively test the JPM with guidance system, we created an e-commerce store environment where the robot receives commands to pick an item from the shelf and place it in the shopping cart. Here, it is crucial to select the correct item exactly as specified in the instructions, so we conducted testing on in-domain and out-of-domain data with new items. We report metrics in \autoref{fig:jpm_res}. 

\begin{figure*}[ht!]
    \centering
    \includegraphics[width=0.9\textwidth]{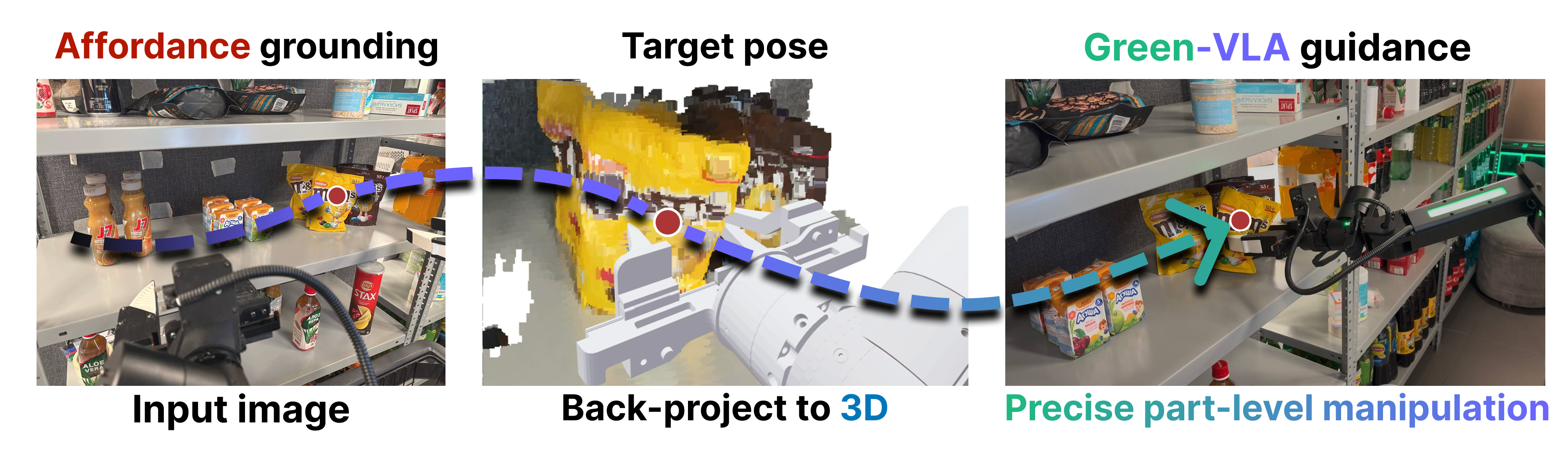}
    \caption{First, we ground the affordance in 2D, and then, by lifting it to 3D, we initialize the target for Green-VLA guidance.}
    \label{fig:jpm_guidance_pipeline}
\end{figure*}

Notably, this e-commerce task is hard for end-to-end VLM policies: they must disambiguate near-identical products (e.g., orange vs. pineapple juice, different apple varieties) where the only reliable cues are small label text, subtle branding, or fine-grained color accents, and packaging often changes, creating OOD variants and near-duplicates. Our JPM predicts a language-conditioned target point, and the guidance module steers flow matching toward that target, yielding a significant success rate boost while preserving policy flexibility. We evaluate shelf picking under three recognition regimes using the same scene layouts for all policies. In in domain (ID)–Coarse, the instruction specifies only a brand/category (e.g., “J7 juice”), and any variant in that class is counted as correct; this measures basic category following under clutter.

In In-D omain Stock Keeping Unit ID–SKU (Fine), the instruction targets an exact variant/SKU (e.g., “J7 orange 0.5 L”), and success requires grasping and placing that precise item—penalizing near-misses like the wrong flavor or size. In out-of-distribution (OOD) evaluation, we introduce unseen SKUs/packaging (new flavors, sizes, or rebranded labels) and assess generalization. For each regime, we randomize object poses and distractors, issue language instructions, and report first-pick correctness (top-1), task success rate (grasp → place), and time-to-completion; mispicks and regrasps are recorded to diagnose failure modes.

\paragraph{Humanoid}

For humanoid evaluation on our Green robot, we focus on instruction-conditioned pick-and-place ensuring robustness to out-of-distribution scene layouts. The humanoid is tasked with picking diverse items — primarily packaged food, the SberBoom smart speaker, and the SberRing smart ring — and sorting them into one of three target baskets according to natural-language commands. 

\begin{figure}[H]
    \centering
    \includegraphics[width=\linewidth]{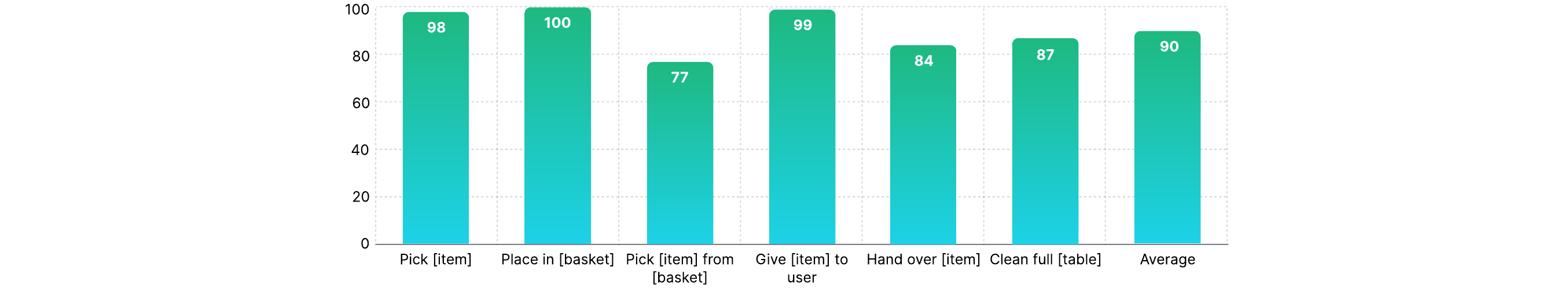}
    \caption{Quantitative evaluation of the humanoid policy. 
    The figure summarizes performance across instruction-conditioned manipulation tasks, including pick, place, handover, fruit sorting, and full table cleaning. 
    The final group shows average success across all tasks, for both in-domain and out-of-domain settings.}
    \label{fig:hum_eval_bench}
\end{figure}

The evaluation includes both left- and right-hand picking, correct basket selection, arm-to-arm handover when required for reachability, and handing objects directly to a user upon request. We randomize object positions, distractor items, and background clutter across episodes to test stability under unseen configurations, and measure success by exact task following: using the instructed arm, selecting the correct object, executing the correct placement or handover, and maintaining safe, reliable behavior in OOD setups. We show several representative tasks captured from the robot’s onboard and external cameras, while the control policy operates the full upper body, including the head, arms, and torso.
% On \autoref{fig:hum_example} we provide several example tasks from the robot’s onboard camera and an external camera setup. We control through policy full upper body (head, arms, torso).

\begin{figure}[H]
    \centering
    \includegraphics[width=0.9\linewidth]{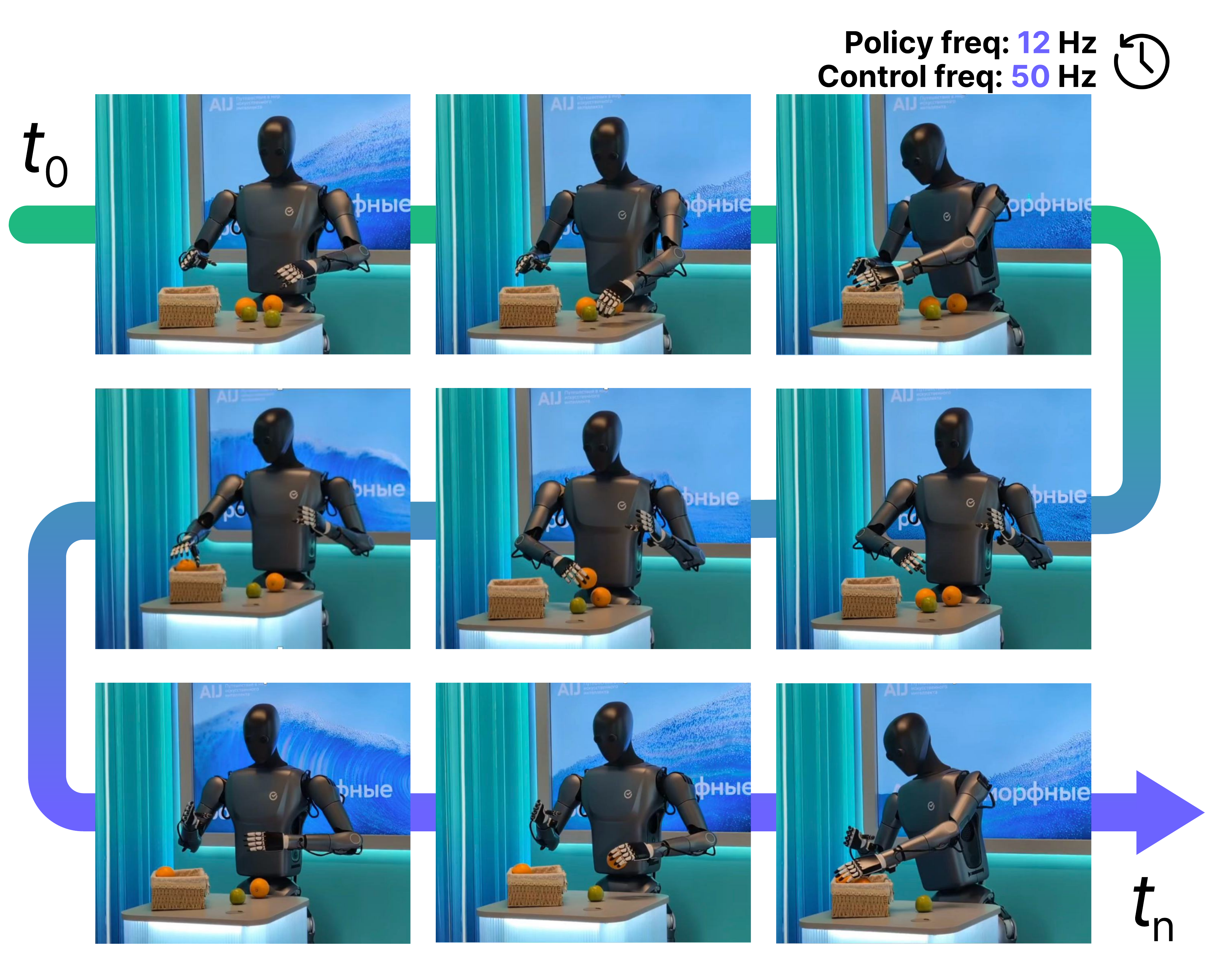}
    \caption{Humanoid executing a task-planned pick-and-place sequence. 
    The high-level task planner decomposes the user's request (“sort apples and oranges into the basket”) into subtasks such as “pick the small green apple with your left hand,” “pick the large orange with your right hand,” and “place the big apple in the target basket.” 
    Green-VLA then executes each subtask, coordinating arm choice, precise placement, full upper-body control, and task-following to complete the full sorting task.}
    \label{fig:hum_sort_ep}
\end{figure}

In the main humanoid scenario, we evaluate the system along several core capabilities. 
First, we test whether the system reliably picks the correct item, in both in-distribution and out-of-distribution scenes, and report average results. For the humanoid scenario we evaluate a set of instruction-conditioned capabilities that together characterize full-table interaction. Each episode is controlled by specifying which hand to use and providing explicit object and basket references. Concretely, we measure:
(i) Pick success for commands of the form pick [item] from the [table/highground] with [left/right] hand;
(ii) Place success for place [item] into the [basket] with [left/right] hand;
(iii) Basket retrieval for pick [item] from the [basket] with [left/right] hand;
(iv) Hand-over to user for give [item] to user;
(v) Fruit sorting, where the robot must pick individual fruits from a central basket and place them into the correct baskets based on language instructions;
(vi) Average chain length (ACL), defined as the number of primitive actions in these composite sequences (pick → place → re-pick → give, up to length 4), which captures how fast the robot completes multi-step tasks; and
(vii) Table cleaning time, the average time to clear all relevant objects via repeated pick–place subtasks, serving as an overall efficiency metric that reflects how quickly the system completes long-horizon workflows that require repeated pick–place subtasks. 

Taken together, these metrics show that the humanoid system can robustly follow detailed language instructions, coordinate both arms, and complete complex sorting and cleaning tasks in a direct and time-efficient manner, even under OOD scene layouts.

\subsection{R2: RL Alignment}

We evaluate the effectiveness of the R2 RL alignment phase on the Simpler BRIDGE WidowX and CALVIN ABC→D benchmarks, along with pick-from-shelf tests in our e-commerce environment.

\begin{figure*}
    \centering
    \begin{subfigure}[t]{0.59\linewidth}
        \centering
        \includegraphics[width=\linewidth]{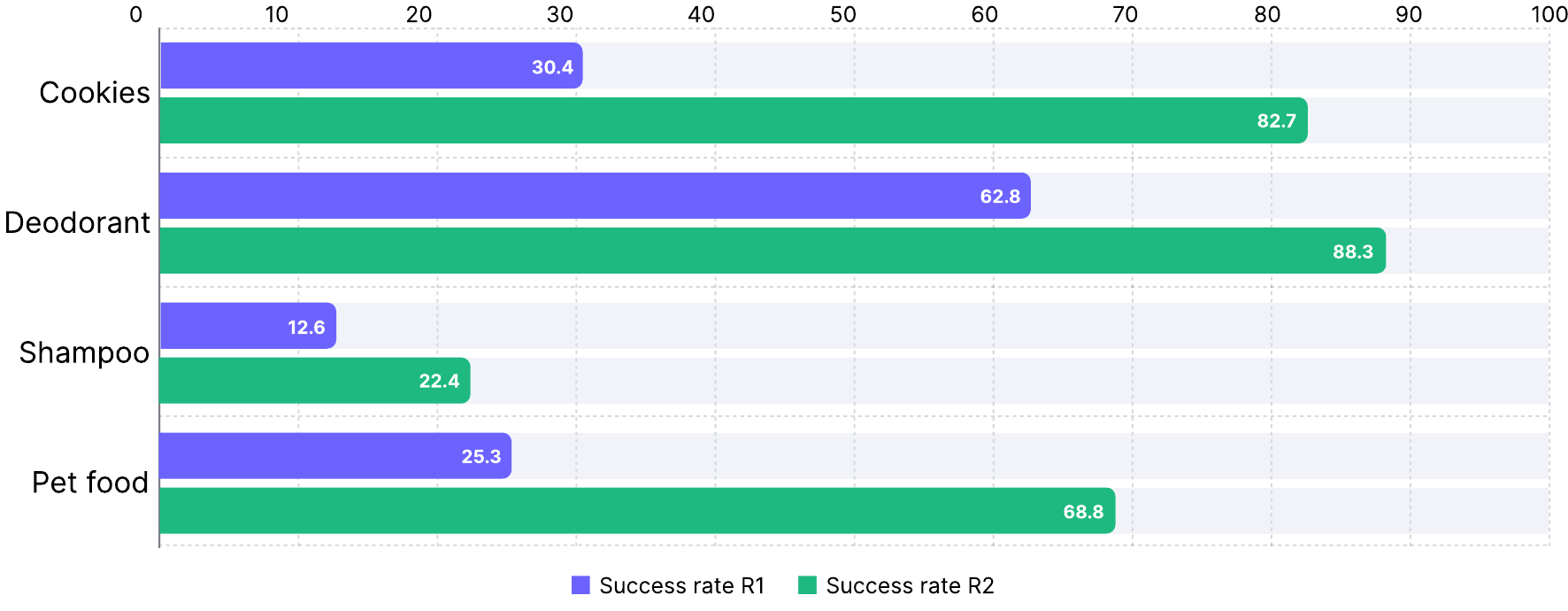}
        \caption{Success rates for e-commerce items after R1 and R2 training stages.}
        \label{fig:r1_r2_ecom}
    \end{subfigure}
    \hfill
    \begin{subfigure}[t]{0.4\linewidth}
        \centering
        \includegraphics[width=\linewidth]{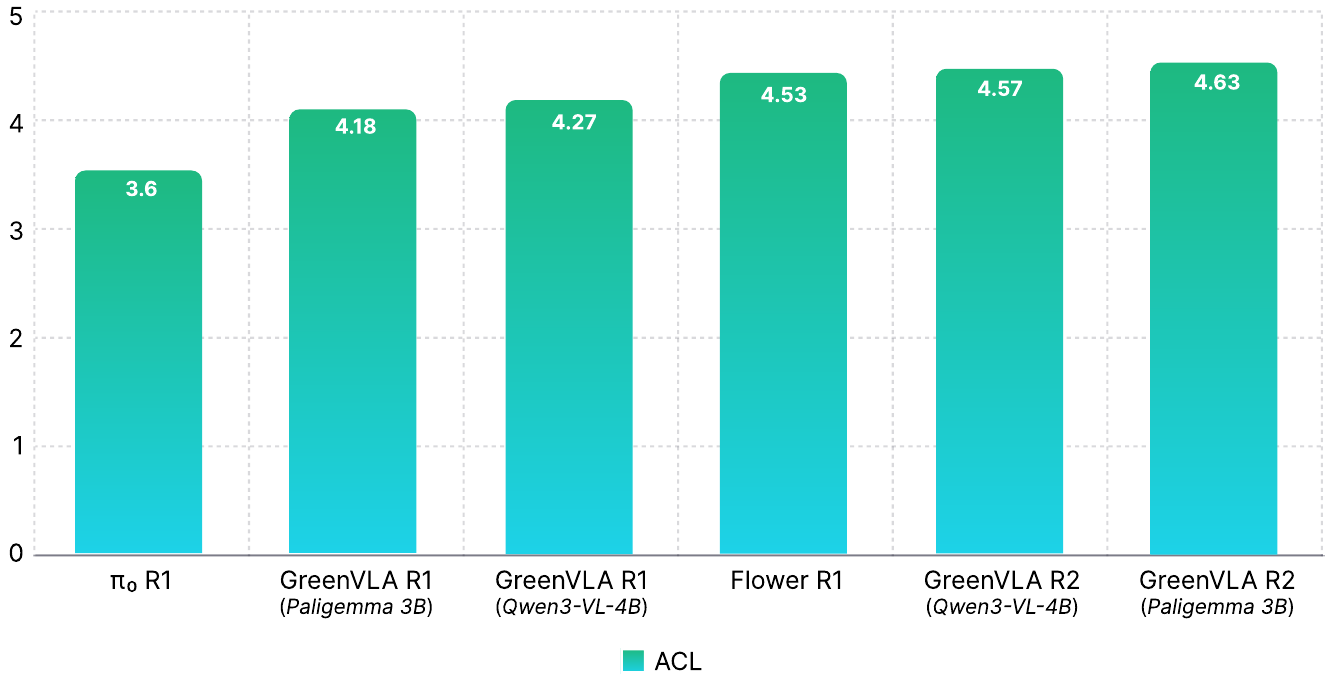}
        \caption{CALVIN benchmark: ACL for $\pi_0$, Flower and Green-VLA across R1/R2.}
        \label{fig:calvin}
    \end{subfigure}
    \caption{Comparison of Green-VLA performance in e-commerce setup and CALVIN ABC$\!\rightarrow\!$D benchmark.}
    \label{fig:overall}
\end{figure*}

On the CALVIN benchmark, we compare three models: (i) $\pi_0$ fine-tuned on CALVIN, (ii) Green-VLA after R1 embodiment fine-tuning on CALVIN, and (iii) Green-VLA after R2 RL alignment. We find that Green-VLA R1 and $\pi_0$ achieve comparable performance, with R1 slightly ahead on aggregate success rate and multi-step task chains. The largest gains come from R2 (see \autoref{fig:calvin}): RL alignment markedly improves long-horizon consistency, error recovery, and compositional task success—raising average chain length (ACL). Overall, reward-shaped refinement helps overcome BC saturation and delivers a substantive performance improvement over both $\pi_0$ fine-tuning and Green-VLA R1. For Green-VLA, we do not use the unified action space for this benchmark.

We apply the same RL alignment procedure in the Simpler BRIDGE WidowX environments and report success rates for the Green-VLA R2 checkpoint in \autoref{tab:fractal-bridge}. R2 alignment improves the R1 model’s success rate by an absolute 24\%, demonstrating high effectiveness of RL fine-tuning.

Finally, we show results of R2 fine-tuning for the “pick and place into tote” task in our e-commerce environment in \autoref{fig:r1_r2_ecom}. While JPM and guidance modules are primarily focused on task following in this setup, R2 RL fine-tuning targets physically reliable grasping of challenging objects. We evaluate on a set of items with difficult shapes, textures, sizes, and weights (e.g., cookies, shampoo), deformable packaging (pet food), and medium-difficulty rigid items (e.g., deodorant). R2 optimization of success-conditioned rewards and penalty-free recoveries leads to more accurate approach trajectories, improved contact geometry, and fewer drops or slips.

\section{Conclusion}

We presented \emph{Green-VLA}, a staged vision–language–action framework that moves beyond raw scale toward quality alignment, action unification, and reinforcement learning fine-tuning. At the data level, our \dataqa{} pipeline filters and smooths heterogeneous demonstrations and aligns temporal dynamics via optical flow–based resampling. At the policy level, a unified action space with embodiment prompts resolves cross-robot inconsistencies and enables positive transfer. At the training level, a target-balanced sampling schedule stabilizes multi-embodiment flow matching, while conservative RL fine-tuning boosts performance on difficult, long-horizon tasks requiring advanced dexterity. Finally, at inference, efficiency optimizations and guidance enable low-latency, instruction-following control—even for novel, language-specified items.

Empirically, Green-VLA demonstrates strong pretrain-stage performance on Simpler and CALVIN, outperforming prior foundation policies at comparable stages and approaching fine-tuned baselines. On real robots, we observe successful application on bimanual setups, and reliable humanoid behavior under OOD layouts. With the R2 RL alignment phase, Green-VLA achieves state-of-the-art results on the Simpler BRIDGE WidowX setup and competitive, near-state-of-the-art performance on CALVIN ABC→D. 

While promising, Green-VLA’s performance still depends on retargeting fidelity, residual dataset bias, and adequate coverage of dexterous skills. Future work will extend multilingual instruction following, strengthen the coupling between fast reasoning and real-time control, and integrate online data collection with safety-aware RL to further reduce failure modes. 

Overall, Green-VLA offers a practical recipe—from web-scale grounding to unified robotics pretraining, embodiment adaptation, and RL alignment—for building generalist, responsive, and reliable robot policies.

\section{Contributors and Acknowledgments}
\label{sec:contributors}
\paragraph{Contributors*:}

\textit{VLA}: I. Apanasevich, M. Artemyev, R. Babakyan, P. Fedotova, D. Grankin, E. Kupryashin, A. Misailidi, D. Nerus, A. Nutalapati, G. Sidorov

\textit{RL fine-tune}: I. Efremov, M. Gerasyov, D. Pikurov, Y. Senchenko

\textit{Data pipeline}: S. Davidenko, D. Kulikov, M. Sultankin

\textit{Control}: K. Askarbek, O. Shamanin, D. Statovoy, E. Zalyaev, I. Zorin % кто-то из команды бади?

\textit{Data collection}: A. Letkin, E. Rusakov, A. Silchenko, V. Vorobyov

\textit{Benchmarks}: S .Sobolnikov

\textit{Project supervisor}:
A. Postnikov

{\scriptsize *Authors are listed in alphabetical order.}

\paragraph{Acknowledgments:}

We sincerely thank Sber and the Sber Robotics Center for the opportunity to work on such an ambitious project. We are especially grateful to the Body, Brain, Platform, Dev, Managers, and People Support teams for their professionalism, continuous support, and close collaboration—making it possible to bring to life what we once only envisioned.

%\clearpage
\bibliographystyle{abbrvnat}
\bibliography{svla}

\end{document}